\documentclass[nonacm]{acmart}

\AtBeginDocument{%
  \providecommand\BibTeX{{%
    \normalfont B\kern-0.5em{\scshape i\kern-0.25em b}\kern-0.8em\TeX}}}

\setcopyright{none}

\usepackage{tabularx}

\begin{document}

\title{Explainable Interface for Human-Autonomy Teaming: A Survey}
\author{Xiangqi Kong}
\email{xiangqi.kong@cranfield.ac.uk}
\author{Yang Xing}
\authornote{Corresponding author}
\email{yang.x@cranfield.ac.uk}
\orcid{https://orcid.org/0000-0002-3786-2865}
\author{Antonios Tsourdos}
\email{a.tsourdos@cranfield.ac.uk}
\author{Ziyue Wang}
\email{ziyue.wang.907@cranfield.ac.uk}
\author{Weisi Guo}
\email{weisi.guo@cranfield.ac.uk}
\author{Adolfo Perrusquia}
\email{adolfo.perrusquia-guzman@cranfield.ac.uk}
\affiliation{%
  \institution{Cranfield University}
  \city{Bedfordshire}
  \country{UK}
  \postcode{MK43 0AL}
}

\author{Andreas Wikander}
\affiliation{%
  \institution{SAAB Group}
  \city{Göteborg}
  \country{Sweden}}
\email{andreas.wikander@saabgroup.com}

\renewcommand{\shortauthors}{XQ.Kong and Y.Xing et al.}

\begin{abstract}
  Nowadays, large-scale foundation models are being increasingly integrated into numerous safety-critical applications, including human-autonomy teaming (HAT) within transportation, medical, and defence domains. Consequently, the inherent 'black-box' nature of these sophisticated deep neural networks heightens the significance of fostering mutual understanding and trust between humans and autonomous systems. To tackle the transparency challenges in HAT, this paper conducts a thoughtful study on the underexplored domain of Explainable Interface (EI) in HAT systems from a human-centric perspective, thereby enriching the existing body of research in Explainable Artificial Intelligence (XAI). We explore the design, development, and evaluation of EI within XAI-enhanced HAT systems. To do so, we first clarify the distinctions between these concepts: EI, explanations and model explainability, aiming to provide researchers and practitioners with a structured understanding. Second, we contribute to a novel framework for EI, addressing the unique challenges in HAT. Last, our summarized evaluation framework for ongoing EI offers a holistic perspective, encompassing model performance, human-centered factors, and group task objectives. Based on extensive surveys across XAI, HAT, psychology, and Human-Computer Interaction (HCI), this review offers multiple novel insights into incorporating XAI into HAT systems and outlines future directions.
\end{abstract}

\ccsdesc[500]{Computing methodologies~Knowledge representation and reasoning}
\ccsdesc[500]{Human-centered computing~Collaborative and social computing}

\keywords{Explainable artificial intelligence, explainability, interpretability, human-autonomy teaming, human-in-the-loop, explainable interface.}

\maketitle

\section{Introduction}
\subsection{Background}
Over the past decade, advancements in machine learning, computing, and communication, have significantly increased the machine autonomy \cite{xing2021humanvehicle}. This evolution from automated to autonomous systems signifies their transition from mere tools to teammates in Human-Autonomy Teaming (HAT). However, the increase in machine autonomy brings challenges in ethics and value alignment, authority transfer explanation, and supervision complexities \cite{li2023trustworthy,methnani2024who}. 

Essential to addressing these challenges is the need for mutual explanation, understanding, and trust, particularly regarding the machine's capabilities in perception, processing, decision-making, and planning in uncertain situations \cite{klien2004ten,azevedo2017vision,Omeiza_2022}. Klein et al. \cite{klien2004ten} emphasize that effective HAT requires four key commitments: team agreement, predictable mutual actions, directability, and maintaining common ground. They also identify ten crucial challenges for creating intelligent and efficient HAT, ranging from adherence to agreements and predictability to shared goal setting, collaborative planning, and optimizing energy in joint activities. Joseph et al. \cite{lyons2021human} identify three key challenges in HAT: creating team affordances for shared awareness and motivation, understanding beneficial tasks and interactions for social cueing, and developing techniques to use these cues to improve HAT performance.

Mutual understanding within HAT systems relies on transparency, accountability, and predictability. Accountability in HAT, as defined \cite{ferber1999multiagent} is "process of account to some agent for a specific task". While related to responsibility and liability, Mulgan \cite{mulgan2000accountability} differentiates them by highlighting that accountability demands justifications for actions and stands indivisible, whereas liability pertains to legal or financial obligations. 

Mutual trust in HAT systems is closely tied to clarity in accountability. Trust grows when users witness autonomous systems performing tasks with reliability and transparency, particularly in elucidating decisions in uncertain scenarios. Minimizing uncertainty and improving transparency in learning are crucial for trust development. This must be coupled with an understanding of human capabilities to optimize task collaboration \cite{wright2020agent,siau2018building}.

To build efficient HAT for time-critical and safe-critical tasks leveraging AI, explainability is central for fostering mutual understanding and trust \cite{alzetta2020intime,madhav2023explainable}. Without two-way explanations between humans and autonomy, the latter becomes a "black box", severing a vital communication link and threatening the collective objective \cite{yuan2022situ}. Humans need to clearly convey their motivations and intentions to machines for goal alignment, while machines should provide clear explanations of their actions.

In conclusion, Explainable Interface (EI) is crucial in HAT for effective collaboration. The core benefits of integrating EI with Explainable Artificial Intelligence (XAI) in HAT encompass: 
\begin{itemize}
  \item \textbf{Enhancing understanding:} EI clarifies how autonomy decisions are made, making complex AI processes transparent and accessible.
  \item \textbf{Enhancing effective communication and trust:} EI facilitates clear, two-way communication between humans and machines, essential for building or restoring trust.
  \item \textbf{Facilitating seamless coordination:} By fostering interaction among HAT agents, EI strengthens the overall system, enabling smoother and more intuitive teamwork.
\end{itemize}

\subsection{Contribution}
Table~\ref{tab:surveys} presents an examination of 30 XAI surveys. We reviewed their coverage of HAT, model explainability, Large Language Models (LLMs) for explanation generation, human perspectives, system design paradigm and evaluation framework. Based on the summarized results, we address the frequently reported lack of structured frameworks and reviews for Explainable Interface enhanced Human-Autonomy Teaming (EI-HAT) systems and respond to call for deeper integration of human-centric considerations in XAI system. 

These previous reviews collectively serve as a robust underpinning for the literature synthesis presented in our paper. In contrast to previous surveys, our study takes a step further by considering the design, development, and evaluation of EI, exploring its critical role in HAT and discussing future trends and challenges in creating more intuitive and user-friendly explainable systems. Key contributions of this study include:

\begin{itemize}
  \item \textbf{Clarifying 'explainable interface':} Differentiating between 'explainability', 'explanation', and 'explainable interface', clarifying roles of AI researchers, developers, and designers.
  \item \textbf{Exploring conceptual and design framework for EI in HAT:} Introducing a framework for developing EI for HAT. Distinguish key points of task-oriented and experience-oriented ones.
  \item \textbf{Summarizing model explainability enhancement:} Organizing model explainability enhancement methods in practical way, specially summarizing the ongoing research on LLMs for explanation generation.
  \item \textbf{Proposing evaluation framework for EI-HAT:} Advocating for a comprehensive assessment of EI in HAT by embracing model performance, human-centered, and group performance evaluation methods.
  \item \textbf{Highlighting challenges and directions:} Enumerating ongoing challenges in the integration of EI for HAT.
  \item \textbf{Providing application perspective insights:} Analyzing the evolution and adaptability of EI across healthcare, transportation and digital twin.
\end{itemize}

\begin{table}[htbp]
  \caption{Overview of Relevant Surveys on XAI and HAT, with Main Topic for Contribution}
  \label{tab:surveys}
  \begin{tabular}{ccccccccc}
    \toprule
    Ref & Year & Topic & HAT & Model & LLMs & Human & Design & Evaluation \\ 
    \midrule
    Ours & 2024 & Explainable Interface, HAT & $\blacksquare$ & $\blacksquare$ & $\blacksquare$ & $\blacksquare$ & $\blacksquare$ & $\blacksquare$  \\ 

    \cite{abdul2018trends} & 2018 & XAI Systems, HCI & - & $\blacksquare$ & - & $\blacksquare$  & $\blacksquare$  & -   \\ 
    \cite{dudley2018review} & 2018 & User Interface, Interactive ML & $\blacksquare$ & - & - & $\blacksquare$ & $\blacksquare$ & $\blacksquare$  \\
    \cite{rosenfeld2019explainability} & 2019 & Explainability, Human-Agent Systems & $\blacksquare$ & $\blacksquare$ & - & $\blacksquare$ & $\blacksquare$ & $\blacksquare$  \\
    \cite{guidotti2019survey} & 2019 & XAI & - & $\blacksquare$ & - & - & - & -   \\
    \cite{barredoarrieta2020explainable} & 2020 & XAI, Responsible AI & - & $\blacksquare$ & - & $\blacksquare$ & - & $\blacksquare$  \\
    \cite{Mohseni_2021} & 2021 & XAI Systems & - & $\blacksquare$ & - & $\blacksquare$ & $\blacksquare$ & $\blacksquare$  \\
    \cite{vilone2021notionsa} & 2021 & XAI & - & $\blacksquare$ & -  & -  & -  & $\blacksquare$  \\
    \cite{wells2021explainable} & 2021 & Explainable Reinforcement Learning & $\blacksquare$ & $\blacksquare$ & - & $\blacksquare$ & $\blacksquare$ & $\blacksquare$  \\
    \cite{chou2022counterfactualsa} & 2022 & Counterfactual and Causality & - & $\blacksquare$ & - & $\blacksquare$ & $\blacksquare$ & $\blacksquare$  \\
    \cite{karimi2022survey} & 2022 & Contrastive Explanations & - & $\blacksquare$ & - & - & - & $\blacksquare$  \\
    \cite{messina2022surveya} & 2022 & XAI, Medical Images & - & $\blacksquare$ & - & - & - & $\blacksquare$   \\
    \cite{vouros2022explainable} & 2022 & Explainable Reinforcement Learning & - & $\blacksquare$ & - & - & - & $\blacksquare$ \\
    \cite{zini2022explainability} & 2022 & NLP, Explainability & - & $\blacksquare$ & - & - & - & $\blacksquare$  \\
    \cite{Omeiza_2022} & 2022 & XAI, Autonomous Driving & $\blacksquare$ & $\blacksquare$ & - & $\blacksquare$ & $\blacksquare$ & $\blacksquare$  \\
    \cite{kaur2022trustworthy} & 2022 & Trustworthy AI & - & $\blacksquare$ & - & $\blacksquare$ & - & $\blacksquare$  \\
    \cite{dwivedi2023explainable} & 2023 & XAI & - & $\blacksquare$ & - & - & $\blacksquare$ & $\blacksquare$  \\
    \cite{han2023interpreting} & 2023 &  Interpreting Adversarial Examples & - & $\blacksquare$ & - & - & - & -  \\
    \cite{Ibrahim_2023} & 2023 & Explainable CNNs & - & $\blacksquare$ & - & - & - & $\blacksquare$  \\
    \cite{li2023trustworthy} & 2023 & Trustworthy AI & - & $\blacksquare$ & - & $\blacksquare$ & $\blacksquare$ & $\blacksquare$  \\
    \cite{Milani_2023} & 2023 & Explainable Reinforcement Learning & - & $\blacksquare$ & - & - & - & $\blacksquare$  \\
    \cite{Nauta_2023} & 2023 & XAI Evaluation & - & $\blacksquare$ & - & - & - & $\blacksquare$  \\
    \cite{Patr_cio_2023} & 2023 & XAI, Medical Images & - & $\blacksquare$ & - & - & - & $\blacksquare$  \\
    \cite{Prado_Romero_2023} & 2023 & Graph Counterfactual Explanations & - & $\blacksquare$ & - & - & - & $\blacksquare$  \\
    \cite{sado2023explainable} & 2023 & Explainable Goal-Driven Agents & $\blacksquare$ & $\blacksquare$ & - & $\blacksquare$ & $\blacksquare$ & $\blacksquare$  \\
    \cite{albahri2023systematic} & 2023 & XAI, Healthcare & - & $\blacksquare$ & - & $\blacksquare$ & $\blacksquare$ & $\blacksquare$  \\
    \cite{ali2023explainable} & 2023 & XAI, Concepts, Challenges & - & $\blacksquare$ & - & $\blacksquare$ & $\blacksquare$ & $\blacksquare$  \\
    \cite{weber2023explaininga} & 2023 & XAI Model Improvement, Challenges & - & $\blacksquare$ & - & - & - & $\blacksquare$  \\
    \cite{gao2024going} & 2024 & Explanation-Guided Learning & - & $\blacksquare$ & - & $\blacksquare$ & - & $\blacksquare$  \\
    \cite{methnani2024who} & 2024 & Trustworthy AI, Variable Autonomy & $\blacksquare$ & $\blacksquare$ & - & $\blacksquare$ & $\blacksquare$ & - \\
    \cite{wu2024usable} & 2024 & XAI in LLM Era & - & $\blacksquare$  & $\blacksquare$ & $\blacksquare$ & $\blacksquare$ & $\blacksquare$  \\

    \bottomrule
  \end{tabular}
\end{table}

\subsection{Paper Organization}
\begin{figure}[htbp]
  \centering
  \includegraphics[width=\linewidth]{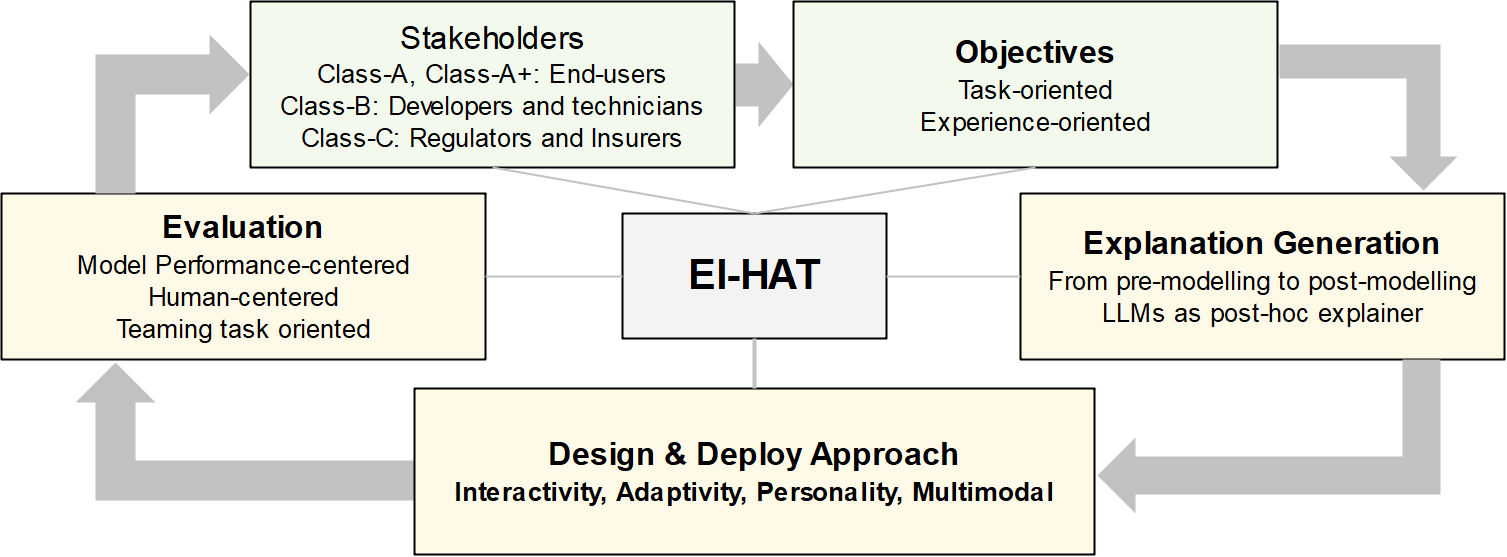}
  \caption{Design Approach for Explainable Interface enhanced Human-Autonomy Teaming.}
  \label{fig:design-framework}
  \Description{Design approch for explainable interface enhanced human-autonomy teaming (EI-HAT). Five key considerations are identified: stakeholders, objectives, explanation generation, design approach and evaluation.}
\end{figure}

As outlines in Figure~\ref{fig:design-framework}, the design approach for EI-HAT system begins with stakeholder requirements analysis. This is followed by defining the system's objectives, which can be task-oriented, experience-oriented, or both. Next is the explanation generation, laying the groundwork for a design approach that encompasses interactivity, adaptivity, personality, and multi-modality. The final phase is system evaluation, which can be model-based, human-centered, or teaming task-oriented. This completes a full cycle of key elements essential for the functionality of the EI-HAT system.

Our paper is structured as follows. In Section~\ref{sec:highlevel}, we present the high-level architecture of EI-HAT, laying the foundational understanding of what the EI is, why it is needed, and how it functions. Building on this, Section~\ref{sec:model} delves into the specific frameworks and techniques for model explainability across various stages, from pre-modeling to post-modeling stage, specially LLMs for explanation generation is discussed. Section~\ref{sec:human} explores the human-centered EI-HAT. To assess the effectiveness of EIs, Section~\ref{sec:evaluation} provides a comprehensive summary of evaluation framework. In Section~\ref{sec:challenges}, challenges and directions are discussed. The study is concluded in Section~\ref{sec:conclusion}. 

\section{High-Level Architectures of Explainable Interface in HAT}\label{sec:highlevel}
\subsection{Clarification of Explainable Interface}
In analysis of the surveys presented in Table~\ref{tab:surveys}, we observed frequent discussions around various XAI-related concepts but still lack of consistency in terminology usage.

BJ Fogg's Prominence-Interpretation Theory, proposed in 2003, can be viewed as an early exploration of the concept of an "Explainable Interface" \cite{fogg2003prominenceinterpretation}. It consists of two main components: Prominence and Interpretation. Prominence pertains to the noticeable and distinct elements on a website that capture user attention. Interpretation involves how these elements are judged by users. For example, a colorful, animated banner ad is prominent and may lead users to judge the site's credibility . The term "Explainable Interface" first emerged in the ACM archives in 2010 \cite{leichtenstern2010managing} in research on the role of trust in self-adaptive ubiquitous computing systems, focusing on the pervasive display environment. It revealed that elements like transparency, controllability, security, privacy, and seriousness positively impact user trust development, while user discomfort, uncertainty, and annoyance negatively correlate with it. The Defense Advanced Research Projects Agency (DARPA)'s XAI program presents it as a two-stage process, separating the XAI model from the user interface for explanations \cite{Gunning_2021}. Study \cite{abdul2018trends} calls for in-depth research and application from professionals in HCI and cognitive sciences. The studies \cite{Mohseni_2021,wang2019designing,dominguez2019effect} discuss EI as a layer presenting reorganized explanatory information, facilitating interaction between humans and autonomy, though they don't provide explicit definitions.

In our paper, we define the concept of an "Explainable Interface" within the context of AI-enhanced HAT systems, distinguishing it from conventional software systems where explanations primarily serve as user guidance. The "Explainable Interface"  is designed to enhance two-way communication between humans and autonomy by reasoning, communicating and adapting to the key working states from both sides.

\begin{figure}[htbp]
  \centering
  \includegraphics[width=\linewidth]{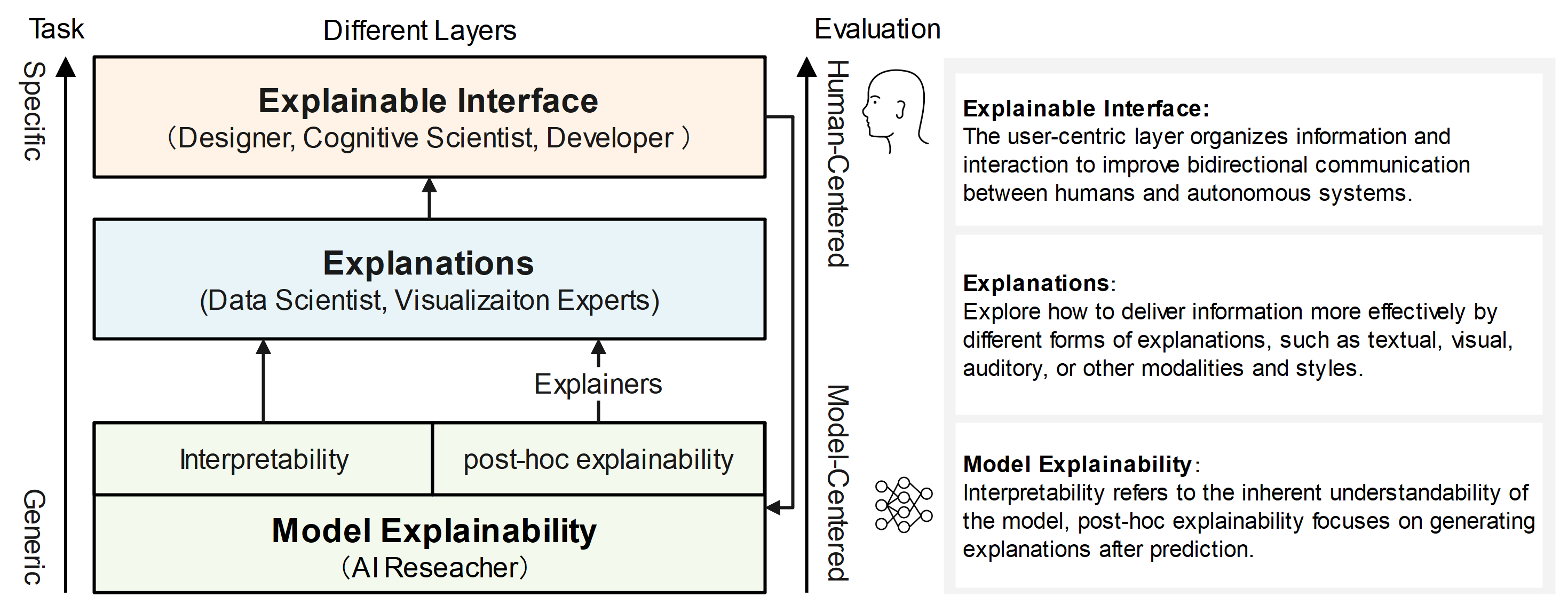}
  \caption{Clarification of Model Explainability, Explanations, Explainable Interfaces.}
  \label{fig:clarification}
  \Description{The concepts of model explainability, explanations and explainable interface, are clarified in this image, progressing from bottom to top. It highlights the increasing importance of task-specific design and human involvement in the design and validation process.}
\end{figure}

As shown in Figure~\ref{fig:clarification}, from bottom to top, \textbf{Model Explainability} concerns itself with two primary facets: the inherent interpretability of AI models and the enhancement of understanding through post-hoc explainability. Interpretability and transparency can help form explanations but are only part of the process \cite{rosenfeld2019explainability}. The metrics of success here include the complexity of the model and the accuracy, completeness, robustness and efficiency of the generated explanations. Such endeavors necessitate the dedicated efforts of AI algorithm researchers. \textbf{Explanations} serve as the information communication with users by different forms of explanations considering their influence of human reasoning\cite{szymanski2021visual}. This necessitates collaborative efforts from experts in data scientists to translate computational outcomes into multimodal information such as graph, image, text, speech \cite{lee2023investigating}. Building upon the foundation of Explainability and Explanations, \textbf{Explainable Interface} addresses specific domain tasks by customizing the presentation of complex AI processes and outcomes in user-understandable and actionable formats, promoting effective HCI through organized multimodal presentations. Achieving this requires expertise from designers\cite{zhu2018explainable}, developers, and cognitive scientists versed in AI and specific domains. Starting from the bottom, tasks become increasingly specific and evaluations shift from focusing on the model to being more human-centered. Clarifying these definitions assists researchers in deconstructing the complexity of XAI-related issues arising from its interdisciplinary nature.

\subsection{The Need of Explainable Interface}
Shahin et al. identified three key reasons for employing EI in HAT, analyzed through psychological, sociotechnical, and philosophical perspectives \cite{atakishiyev2023explainable}. Psychologically, EI enhances safety and reduces accidents by leveraging concepts like the Theory of Mind for user-centric explanations \cite{akula2022cxtom}. Sociotechnically, EI is essential for the human-focused design of systems, improving collaboration. Philosophically, it argues that AI systems should proactively provide explanations of their decisions and future strategies, aiding in deeper human understanding.

\subsubsection{Regulation and Guidelines}
DARPA formulated the XAI program in 2015 with the goal to enable end users to better understand, trust, and effectively manage AI systems \cite{Gunning_2021}. General Data Protection Regulation (GDPR) emphasize on clarity in decisions stemming from human data, underscoring the imperative of user comprehension and their right to question system decisions \cite{voigt2017eu}. In 2023, the US National Institute of Standards and Technology (NIST) introduced the AI Risk Management Framework 1.0 (RMF). This framework provides valuable guidance for tech entities engaged in AI, emphasizing the importance of trust and responsibility in implementation. One common misconception is the idea that AI systems are inherently objective and error-free. This misconception can have detrimental effects on individuals and broader ecosystems. To address this, the RMF outlines key pillars of AI trustworthiness, including safety, security, resilience, explainability, privacy, fairness, accountability, transparency, validity, and reliability \cite{ai2023artificial}.

In healthcare, clinical stakeholders must comprehend AI recommendations to build trust and make informed decisions\cite{chen2022explainable}. This becomes even more significant when certain AI models are classified as medical devices by regulatory bodies like the US Food and Drug Administration (FDA), subjecting them to strict regulations \cite{pesapane2018artificial, health2023artificial}. In this situation, effective EIs are essential in areas such as diagnostic support tools, treatment recommendation systems, risk prediction models, and medical image analysis tools.

\subsubsection{Expectation of Different Stakeholders}
When designing EI-HAT system, it is imperative to consider stakeholders which can be categorized by roles and expertise from the start because of their different requirements.

Arrieta et al. \cite{barredoarrieta2020explainable} identify stakeholder categories including domain experts, data scientists, managers, and regulatory entities. Langer et al. expand this to five classes: users, developers, affected parties, deployers, and regulators, noting differing expectations like acceptance and accountability \cite{langer2021what}. Building upon the categories outlined in \cite{Omeiza_2022}, we introduce an additional key category: Class-A+, which includes individuals with high technical proficiency, such as pilots or operators of advanced military autonomous systems. Class-A covers end-users without specialized knowledge, who require user-friendly,  simplified, understandable AI explanations and privacy considerations \cite{jin2023invisiblea,gerlings2022explainable}. Class-A+, for technically adept end-users, demands in-depth, sector-specific AI information \cite{gerlings2022explainable,sheth2021knowledgeintensive}. Class-B, encompassing developers and technicians, concentrates on detailed AI model insights and ethical development approaches, using explanations for model improvement, debugging, and modification \cite{langer2021what,idahl2021benchmarking,gao2024going,Teso_2023}. Class-C includes regulators and insurers, focusing on adherence to standards like GDPR and dependability in critical applications \cite{ehsan2021expanding,voigt2017eu}. Importantly, we recognize developers' dual role: as users needing comprehensive model understanding, and as creators who must consider EI characteristics for their intended audiences.

Moreover, the range of stakeholders in XAI systems differs according to the design domain. In the healthcare sector, for example, key stakeholders include data scientists, clinical researchers, and clinicians \cite{dey2022humancentered,kim2024stakeholder}. Additionally, due to the similarity in target groups, for interfaces primarily aimed at Class-A, some design experiences from business-to-customer can be referenced. For interfaces mainly targeting Class-A+ and Class B, some design experiences from business-to-business products can be useful.

\subsection{Architecture for Explainable Interface in HAT}
The design of EI-HAT systems does not adhere to a one-size-fits-all solution due to the vast diversity in applications, users, scenarios, and environments. It necessitates a case-by-case examination. Despite this, common design principles emphasize human-centered automation \cite{chen2022explainable,panigutti2023codesign,Mohseni_2021,meske2022design}. This approach highlights the importance of human control, requiring human engagement and informed collaboration through effective communication. Automation should only be introduced when necessary and should be transparent, understandable, predictable, and subject to human direction. It is essential for the system to be able to understand its human counterpart, promoting a shared understanding of intentions, situational awareness, actions, and emotions between humans and automated teammates. Furthermore, the design of these systems should prioritize ease of learning and operation, ensuring simplicity and user-friendliness in human interaction.

\begin{figure}[htbp]
  \centering
  \includegraphics[width=\linewidth]{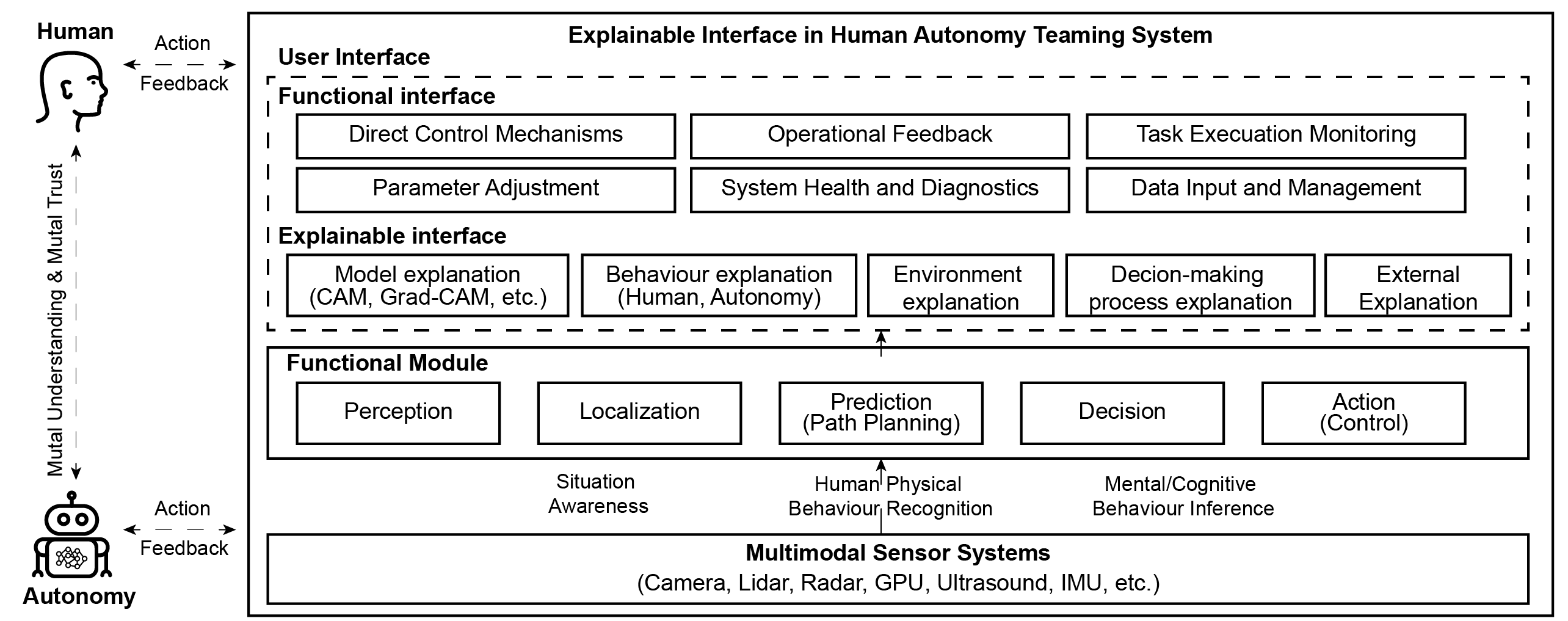}
  \caption{Explainable Interface in Human-Autonomy Teaming System.}
  \label{fig:components}
  \Description{This figure shows the components of explainable interface in human-autonomy teaming system. Explainable interface is seen as a middle layer between human and autonomous agents. It components model explanation, behavior explanation, enviroment explanaiton, decision-making process explanation and external explanation.}
\end{figure}

As depicted in Figure~\ref{fig:components}, The architecture of EI-HAT system consists of user interface, functional module, support systems, and human-autonomy loop. The interface includes functional and explainable components. The functional module comprises perception, localization, prediction, decision, and action \cite{sado2023explainable, xing2023learning}. Support systems include situation awareness, human physical behavior recognition, mental/cognitive behavior inference, and multimodal sensor systems. The human-autonomy loop ensures continuous feedback and promotes mutual understanding and trust.

EI in the architecture includes explanations of models, behaviors, environments, decision-making, and sometimes external explanation \cite{abdul2018trends}. It also gathers feedback from users, while the functional interface continues to focus on direct system interaction and control. EI and functional interface can either be integrated into a single, cohesive user interface or they can be separate, depending on the design and the requirements of the system. There is also a hybrid approach where the core functional interface is supplemented by on-demand explainable features. The design choice should be driven by considerations such as user experience, system complexity, and the cognitive load on the users. Moreover, to ensure the human operator can be effectively explained for that to build proper mutual understanding and mutual trust, human behavior/intention recognition is also required. 

Within such a collaboration system, several negative consequences to the human who work together with the automation can be estimated such as out-of-loop performance, loss of situation awareness, complacency, or lack of vigilance, over-trust, and automation surprises. Accordingly, two different types of assistance from automation can be developed, which are soft protection and hard protection \cite{itoh2014design}. Soft protection in automation helps humans avoid collisions by providing warnings and hazard explanations, and addresses unsafe human behaviors within safety parameters. However, it may not always prevent negative events, especially if humans with inadequate situational awareness override it. In contrast, hard protection actively prevents humans from taking dangerous actions through machine-initiated decisions and controls, which can sometimes lead to unexpected responses from operators unfamiliar with the system. Therefore, there is a need for EIs that allows humans to share their situational awareness with automation, comprehend and interpret its decisions and judgments, and recognize its functional limitations.

\section{Frameworks and Techniques for Model Explainability}\label{sec:model}
Explanations generated across varying phases can serve as input data for EIs. Based on the requirements of different stakeholders, appropriate information can be selected for integration. Explainability methodologies can be categorized into three stages: pre-Modeling Explainability, inherently Interpretable model by Design, and post-Modeling explainability \cite{minh2022explainable}. This categorization facilitates the consideration and selection of explainability methods at various stages of engineering practice. The classification of model explainability methods is diverse, partly due to the multidisciplinary nature of the field. Timo Speith suggests that confusion over taxonomies in the field may delay the potential social benefits of XAI \cite{speith2022review}. 

\begin{figure}[htbp]
  \centering
  \includegraphics[width=\linewidth]{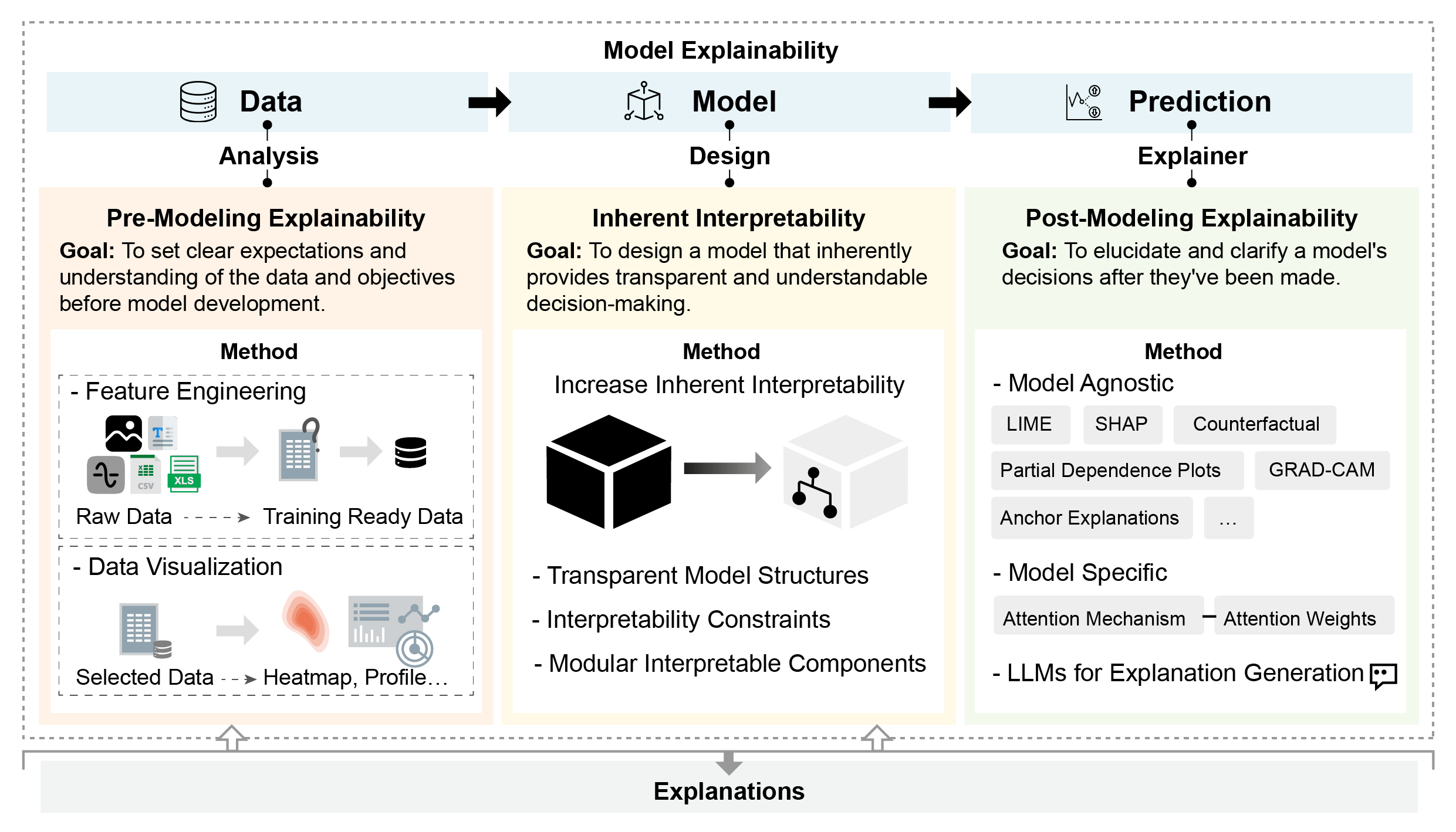}
  \caption{A Comprehensive Framework for Model Explainability in Machine Learning.}
  \label{fig:modeling}
  \Description{This figure outlines a structured approach to explainability throughout the lifecycle of a machine learning model. It's divided into three main stages: Pre-Modeling Explainability, Inherent Interpretability within Model Design, and Post-Modeling Explainability. In the "Pre-Modeling Explainability" phase, the goal is to understand the data and objectives clearly before developing the model. Methods here include feature engineering and data visualization, transforming raw data into training-ready data and visually interpreting selected data via heatmaps and profiles. The "Model Design" stage focuses on "Inherent Interpretability," aiming to create models that are transparent and understandable by design. This includes the creation of globally interpretable models with constraints and model extraction techniques such as linear regression and decision trees, as well as locally interpretable models which may include attention mechanisms and causality enhancement. The final stage, "Post-Modeling Explainability," aims to clarify a model's decisions after they have been made. Methods for this stage are divided into model-agnostic approaches like LIME, SHAP, Counterfactual explanations, Partial Dependence Plots, GRAD-CAM, and Anchor Explanations, and model-specific methods such as attention weights. Additionally, Large Language Models (LLMs) can be employed for explanation generation.}
\end{figure}

As shown in Figure~\ref{fig:modeling}, we have adopted a stages-based approach to minimize overlap and provide practical guidance. In the pre-modeling stage, data analysis is conducted to set clear expectations and understand the data, addressing biases and inconsistencies early on datasets. The model design stage focuses on building inherent interpretability through transparent structures, interpretability constraints, and modular components. Post-modeling methods involve explaining the trained functional models after they made predictions.

\subsection{Pre-modeling Explainability}
Pre-modeling explainability involves data pre-processing and feature exploration for better understanding and preparation of datasets prior to model training, which is crucial for model effectiveness and interpretability \cite{minh2022explainable}. It encompasses two main areas: data visualization and feature engineering. Data visualization utilizes graphical representations, like graphs and heatmap, to reveal data patterns, with advanced methods incorporating ML to enhance design and explanation (e.g., ML4VIS studies) \cite{wang2022survey}. This technique not only helps in identifying data trends but also improves stakeholder understanding and memory retention \cite{waller2020effect}. In sectors such as healthcare, finance, and environmental monitoring, data visualization effectively communicates complex information to non-experts, impacting attitudes and decisions \cite{park2022impact}. Feature engineering involves creating interpretable features through domain knowledge, thus aiding in model decision-making processes \cite{hong2020human}. This includes selecting and transforming data into relevant features to simplify the model and boost its performance \cite{xiang2021physicsconstrained}. The created features should be easily understandable and applicable to the specific domain, like tumor characteristics in medical imaging \cite{chen2022explainable} or word frequency in natural language processing \cite{zini2022explainability}. By incorporating expert input in the feature selection process, the human-in-the-loop approach greatly enhances the accuracy and interpretability of models, particularly in fields such as image analysis and risk assessment \cite{correia2019humanintheloop}. This refinement in selecting relevant features leads to more precise and insightful results. 

\subsection{Inherent Interpretability of Model by Design}
In regulated domains like healthcare and safety, inherent interpretable models are essential due to their clarity and ease of justification during regulatory reviews \cite{sudjianto2021designing}. Post-hoc explanations for black box models may not accurately reflect the original model's computations, or they might be too vague, leading to overly complex decision-making processes prone to errors. Therefore, in high-risk scenarios, inherently interpretable models are more recommended \cite{rudin2019stopa}. These models are designed with a focus on embedding interpretability within their fundamental architecture and optimization processes from the beginning, as opposed to relying on retrospective explanations applied to black box models. There are several design methods to consider when creating inherently interpretable models.

\subsubsection{Using Transparent Model Structures}
Early AI systems, using logic and symbolic reasoning, could perform deductions and provide explanations for their actions. Some commonly used methods include linear regression, decision trees, and rule-based systems.

\begin{itemize}
  \item \textbf{Linear regression:} Statistical method for analyzing the relationship between variables.
  \item \textbf{Decision trees:} Visual decision-making structures that use if-else rules to classify data or make predictions \cite{loh2011classification}. Their interpretability can be quantified through 'explanation size', referring to the attributes assessed from the root to a leaf \cite{souza2022decision}.
  \item \textbf{Rule-based systems:} Transparent systems that deliver decisions based on predefined rules. Highly interpretable, commonly used in expert systems for guidance and recommendations. 
\end{itemize}

One example of an interpretable system is Full Spectrum Command, designed to facilitate training in light infantry command. It allows users to review computer-controlled soldiers' actions and decisions post-mission by clicking on them for detailed explanations\cite{vanlentexplainable}. However, despite its potential, this kind of systems encounter challenges such as inefficiency, high construction costs, and limitations in handling real-world complexity.

\subsubsection{Imposing Interpretability Constraints}
Beyond the architecture of models, imposing specific constraints during model development plays a pivotal role in enhancing inherent interpretability. The following types of constraints are discussed:
\begin{itemize}
  \item \textbf{Enforced sparsity:} This constraint focuses on reducing the complexity of models by limiting the number of active features or parameters. The Sparse Neural Additive Models (SNAM) apply  group sparsity regularization to achieve feature selection and improved interpretability \cite{xu2022sparse}. Yu et al. \cite{yu2023whitebox} propose that representation learning's objective is to compress and transform data distribution into a mixture of low-dimensional Gaussian distributions on incoherent subspaces. They introduce 'sparse rate reduction' as a unified measure for the quality of representations.

  \item \textbf{Directed knowledge-integrated training:} This constraint aligns machine learning models with established principles and causality in a specific field. This approach involves integrating domain-specific knowledge into the training process. Various methods, such as physics-informed neural networks, saliency-guided training, causal knowledge integration, merging knowledge graphs with neural networks, and explanation-guided learning, are used to achieve this alignment \cite{seyyedi2023machine, ismail2021improving, wu2023causality, rozanec2022knowledge, gao2024going}. In healthcare, Domain Knowledge Guided Recurrent Neural Networks (DG-RNN) demonstrate state-of-the-art performance in various healthcare prediction tasks using real-world electronic health record (EHR) data \cite{yin2019domain}.

  \item \textbf{Model streamlining constraints:} This approach encompasses Progressive Mimic Learning (PML) which aims to shape lightweight CNN models by emulating the learning trajectories of more complex models \cite{ma2021progressive}. The use of knowledge distillation, which involves training a smaller student model to mimic the behavior of a larger teacher model achieves a better balance of interpretability and performance \cite{han2023impact}. These strategies  effectively address both the complexity and efficiency aspects in model development.
\end{itemize}

\subsubsection{Combining different interpretable modules} Combining different interpretable modules can have a profound impact on both the performance and interpretability of models. We have explored several approaches that leverage the strengths of interpretable modules to achieve these goals.

\begin{itemize}
  \item \textbf{Generalized Additive Models (GAMs):} GAMs are statistical models that enable flexible, non-linear modeling of relationships between variables. They combine multiple smooth functions to capture the complex patterns in data, with each function representing a different predictor's influence. These models are additive, with the overall output being the sum of these individual functions, and are widely used for their interpretability in fields requiring analysis of complex, non-linear relationships. such as identification of key factors in the success rate of cardiac paralysis surgery \cite{hastie2014generalized}.

  \item \textbf{Neural Additive Models (NAMs):} NAMs blend the principles of interpretable regression components with the power of neural networks. This combination maintains an interpretable structure while benefiting from the expressivity of neural networks \cite{agarwal2021neural}. Recent advancements in NAMs include approaches from a bayesian perspective to enhance transparency, provide calibrated uncertainties, select relevant features, and identify interactions. The development of Laplace-approximated NAMs (LA-NAMs) shows improved performance on tabular datasets and real-world medical tasks\cite{bouchiat2024improving}.

  \item \textbf{Attention Mechanisms:} Attention Mechanisms enable models to focus selectively on relevant parts of input data, assigning varying weights to different input components. This process aids in tracing the model's decision-making back to specific input features. It began with Bahdanau-style attention \cite{bahdanau2016neurala}, which introduced the additive attention, enabling a model to automatically learn to focus on various parts of the input sequence during decoding, thereby enhancing translation performance. This concept was refined by Luong et al. with more effective and computationally efficient multiplicative attention techniques \cite{luong2015effective}. The application of attention mechanisms expanded to image captioning in \cite{xu2016show} where hard attention was applied to selectively concentrate on specific image parts for generating descriptive text. The introduction of transformer architecture, which relies solely on self-attention mechanisms without recurrent or convolutional layers, has drastically improved the quality of machine translations and setting a new standard for sequence-to-sequence models \cite{vaswani2017attention}. The advancements continued with the development of Sparse Transformers, Longformer, and Big Bird, addressing the limitations of standard attention mechanisms in processing long sequences. By employing sparse attention patterns they reduce computational complexity and enable the handling of longer sequences more efficiently \cite{child2019generating,beltagy2020longformer,zaheer2021big}. 
\end{itemize}

In conclusion, it's important to recognize that the methods discussed are not mutually exclusive. For example, \cite{yu2023whitebox} illustrates a synergistic method that combines different techniques, such as unrolling iterative optimization schemes, reinterpreting self-attention layers, and implementing an idealized model for token distribution, all within a cohesive framework aimed at developing transformer-like network architectures. This approach not only enhances the model's mathematical interpretability but also boosts its practical performance. 

\subsection{Post-modeling Explainability}
Post-modeling explainability methods are classified by approach (model agnostic or specific), scope (local or global) \cite{speith2022review}, and the type of output they generate (textual, visual, or quantitative) \cite{vilone2021classification}. The process from pre- to post-modeling explainability is iterative and involves sequential optimization. Three methods of post-hoc explanation are outlined for image and time-series datasets: factual, counterfactual, and semi-factual approaches \cite{kenny2021posthoc}. 

\subsubsection{Feature-based Explanation}
Feature-based explanation is a widely used approach in XAI methods for helping users understand the importance of features. The concept of feature importance is the most prevalent explanation type \cite{Nauta_2023}. Gradient-based methods, such as Grad-CAM \cite{selvaraju2017gradcam} and Integrated gradients \cite{sundararajan2017axiomatic}, compute the gradient of the model's output with respect to its inputs to generate saliency maps. DeepLIFT \cite{shrikumar2019learning} is a method for decomposing the output prediction of a neural network on a specific input by backpropagating the contributions of all neurons in the network to every feature of the input. Perturbation-based methods alter or remove portions of the input to measure their impact on the model's output. Examples include RISE \cite{petsiuk2018rise} and LIME \cite{ribeiro2016why}, which are computationally intensive but applicable to various models. SHAP is a well-known feature-based explanation method that applies Shapley values derived from cooperative game theory to the context of ML. Furthermore, TsSHAP \cite{raykar2023tsshap} targets robust, model-agnostic explainability in univariate time series forecasting.

In addition to analyzing individual feature importance, understanding the interactions between features and their impact is crucial. However, generating feature interactions is more complex compared to assessing individual feature importance. Current research has explored measuring interactions through tree-based models like CART and random forests \cite{lundberg2020local} and visualizing partial dependence plots \cite{inglis2021visualizing}. Nonetheless, further investigation is needed to fully understand the contribution of feature interactions to prediction results. 

\subsubsection{Example-based Explanation}
Example-based methods explain models by selecting instances from the dataset rather than creating feature summaries. Example-based explanations are meaningful only when we can represent data instances in a way that is understandable to humans. It is particularly effective for images, as we can directly view them.

Jacob Bien et al. \cite{bien2011prototype} introduce prototype methods for classification, which aim to select minimal subsets of representative samples to condense the datasets. In \cite{kim2016examples}, Kim et al. highlight the limitations of prototypes in complex distributions. They propose the "MMD-critic" approach, which combines prototypes with a criticism mechanism to enhance human understanding. Jonathan Crabbé et al. \cite{crabbe2022labelfree} address the challenge of interpreting unsupervised black-box models. They introduce label-free feature importance and label-free feature example importance.

\subsubsection{Concept-based Explanation}
Concept-based explanation methods aim to interpret the internal representations or neuron activations of models against high-level concepts. These concepts are characterized by a set of samples that share specific features. These explanations can assist in scientific research by aligning model predictions with known scientific concepts, potentially uncovering new insights.

Anchors technique presents a rule-based, model-agnostic approach for generating precise, user-friendly explanations of predictions \cite{ribeiro2018anchors}. Kim et al. introduced Concept Activation Vectors (CAVs) to interpret neural networks in relation to human-friendly concepts \cite{kim2018interpretability}. CAVs represent the internal states of a neural network as vectors indicating a concept's direction in the model's latent space. Furthermore, Zhang et al. \cite{zhang2021invertible} introduced an Invertible Concept-based Explanation (ICE) framework to enhance concept-based explanations. The framework utilizes non-negative matrix factorization to generate Non-negative Concept Activation Vectors (NCAVs). Human subject experiments demonstrated that NCAVs outperformed previous methods such as PCA. In 2022, CARs is proposed as an extension. CARs represent concepts as regions rather than fixed directions in the Deep Neural Network (DNN)'s latent space, which enable it capture more complex relationships between concepts and the DNN's internal representations \cite{crabbe2022concept}.

In summary, feature-based methods prioritize individual data attributes, example-based methods focus on specific instances, and concept-based methods connect decisions to human-understandable concepts. These methods are often model-agnostic, allowing different explainers to be used with a functional model. The selection of an explainer should consider the domain, AI task, and stakeholder requirements. Additionally, interactive tools can enhance the selection process and improve the XAI process.

\subsection{Causal Inference and Counterfactual Explanation}
Causality and counterfactual reasoning are key in improving mental models by establishing causal connections or directing focus in HAT dynamics \cite{andrews2023role}. The D-BIAS system exemplifies this with a causality-based human-in-the-loop framework to mitigate algorithmic bias in HAT \cite{ghai2022dbiasa}. Similarly, the trust-POMDP model highlights causality's role in HCI, enabling robots to gauge human trust, understand their actions' causal effects on trust, and choose trust-enhancing actions for team success. Importantly, this model reveals that prioritizing team trust, sometimes a counterfactual objective, may not always coincide with maximizing team performance \cite{chen2018trustaware}.

Traditional Causal inference is a field origin in 20th century in the research of statistics and medicine that dedicated to understanding and quantify the causal relationships between variables \cite{pearl2009causal}. Causal inference involves discerning causal relationships between features and variables via causal learning. This entails the extraction of high-level causal variables from low-level observations, thereby facilitating a comprehensive understanding of the underlying causal mechanisms \cite{scholkopf2021causal}. Causal learning is important in ethical AI for actionable healthcare \cite{prosperi2020causal}. 

By modeling causal relationships, machines can better support counterfactual inference therefore be more effective at modifying their behaviors to engage in a truly adaptive team \cite{stowers2021improving}. By incorporating causal relationships, which encompass valuable human knowledge, into the structural design, users can gain deeper insights into the decision-making processes of complex deep learning models, thereby enhancing their interpretability \cite{wu2023causality,scholkopf2021causal}. 

Judea Pearl has proposed a theoretical framework known as the "ladder of causality," which elucidates three distinct levels of causal inference: association, intervention, and counterfactual \cite{pearl2018book}. Using counterfactual to create feature attribution explanations from hypothetical scenarios allows for the identification of key features influencing model classifications. This approach offers a deeper analysis than attribution-only methods by determining which features are necessary or sufficient for altering outcomes \cite{kommiyamothilal2021unifying}. Psychological field also have demonstrated that counterfactual can provide psychologically intuitive and plausible explanations, further affirming their utility in fostering a comprehensive understanding \cite{keane2021if}. Sandra Wachter et al. recommend utilizing the approach of counterfactual explanations without opening the black box to achieve explainability. This approach is based on three explanatory objectives: (1) informing and aiding individuals in understanding why specific decisions were made, (2) providing reasons for questioning those decisions by presenting alternative outcomes, and (3) understanding the necessary changes that need to be made to achieve desired results based on the current decision model \cite{wachter2017counterfactual}.

The Recent studies show a merging of neuro-symbolic methods in causal inference and counterfactual explanation \cite{wu2023weakly,mao2019neurosymbolic}, emphasizing the importance of domain knowledge. However, experiments on Amazon Mechanical Turk reveal that the influence of causality on decision-making varies with individual domain knowledge and doesn't always enhance decision support \cite{zheng2020how}. Counterfactual explanations face limitations including lack of user studies, defining plausibility, sparsity issues, coverage assessment, and comparative testing \cite{keane2021if}. Research \cite{chou2022counterfactualsa} indicates that existing model-agnostic counterfactual algorithms for XAI lack a causal theoretical foundation, thus limiting their ability to convey causality to decision-makers.

\subsection{Explainable Reinforcement Learning}
Explainable Reinforcement Learning (XRL) presents unique challenges in creating and presenting explanations. These arise from its sequential, temporal, and stochastic nature, as well as the complexity of its policies and state spaces \cite{huber2023ganterfactualrl}. This distinctiveness positions XRL as a particularly challenging branch of XAI. 

XRL seeks to improve human collaboration, visualization, policy summarization, query-based explanations, and verification, but faces issues like limited detail in human experiments, scalability, and the complexity of explanations for non-experts \cite{wells2021explainable}. Current XRL methods often derive from supervised learning, but there is a need for RL-specific approaches that include feature importance, learning process, Markov Decision Process, and policy-level explanations \cite{Milani_2023}. XRL-Bench benchmark suite provides a standard and unified platform for researches and practitioners to develop, test, and compare state importance-based XRL algorithms \cite{xiong2024xrlbench}. 

Herman Yau et al. argue that traditional post-hoc explanations are impossible in traditional reinforcement learning. They propose collecting explanations during the agent's training process and derive approaches to extract local explanations based on intention. Their method ensures consistency with learned Q-values and focuses on the intended outcomes of the agent. The researchers evaluate their approach on standard environments using OpenAI Gym and show its superiority over vanilla Q-Learning by providing contrastive explanations. Even for deep Q-networks (DQN) without mathematical guarantees, they are able to approximate the agent's intention \cite{yau2020what}. 

Unlike existing methods that require extensive access to an agent's inner workings, Huber et al. \cite{huber2023ganterfactualrl} redefine the generation of counterfactual states as a domain transfer problem, leveraging the StarGAN architecture to transform features relevant to an agent's decision while maintaining other features. This method aims to create counterfactual states that prompt the agent to choose a desired action, thereby offering a more straightforward and effective way to understand and explain RL agent strategies.

Study \cite{kenny2022towards} introduces the Prototype-Wrapper Network (PW-Net), which integrates human-friendly prototypes into the decision-making process of agents. Additionally, the Human-in-the-loop Explainability via Deep Reinforcement Learning (HEX) method aims to generate personalized explanations for individual decision makers \cite{lash2022hex}. Study \cite{druce2021explainable} creates an EI for Amidar video game with six major components: agent selection, agent description, live game stream, agent performance characteristics, narrative explanation, ”what if” scenarios. Results indicate that EI increases user trust and acceptance, though not necessarily perception of prediction accuracy. Building on this, future research could explore the design and evaluation of EIs in scenarios where AI functions as a trusted collaborator in HAT, rather than operating autonomously.

In RL-enhanced systems, by maintaining human-in-the-loop through EIs, additional context information can be provided to improve the learning process. Human interaction and guidance during the learning process aligns AI decisions with human understanding and values, resulting in more explainable and reasonable behavior from AI agents. And In combining XRL in HAT system, future research could explore immersive visualization and symbolic representation integration \cite{wells2021explainable}.

\subsection{LLMs for Explanation Generation}\label{sec:llm}
The development of LLMs such as GPT-4 \cite{achiam2023gpt} and LLaMA \cite{touvron2023llama} represents a significant progression in natural language processing (NLP). The integration of multi-modality into LLMs has expanded traditional NLP, encompassing diverse data formats such as images \cite{wu2024usable}, audio, and video. 

LLMs assist in analyzing specific behaviors of smaller models. The Transformer Debugger \cite{mossing2024tdb}, utilizing automated interpretability methods and sparse autoencoders. Study \cite{zhang2023explaining} proposes an approach to explain agent behavior by distilling the agent's policy into a decision tree, forming a behavior representation from it, and then querying an LLM for explanations based on this representation.

Nicholas Kroeger et al. \cite{kroeger2023are} explore advanced LLMs' utility for post-hoc explanations through in-context learning (ICL) \cite{dong2023survey}. They developed various strategies like perturbation and explanation-based methods, comparing them against standard explainers such as LIME and SHAP. The results highlighted LLMs' potential as effective post-hoc explainers. Additionally, Feldhus et al. \cite{feldhus2023saliency} examines how converting saliency maps into natural language can make AI explanations more understandable, comparing two novel methods (model-free and instruction-based verbalizations) and evaluating their effectiveness through human and automated tests. The study \cite{chandrasekaran2018explanations} focused on human prediction of Visual Question Answering (VQA) model responses after receiving explanations and revealed no significant improvement in team collaboration tasks such as Failure Prediction (FP) and Knowledge Prediction (KP). Another study \cite{rozanec2022knowledge} explored integrating LLMs with Knowledge Graphs (KGs) to enhance model transparency and interpretability in automotive demand forecasting, aiming to mitigate hallucination challenges and improve understanding of LLMs with real-world data and exogenous factors. While enhanced capabilities of LLMs facilitate their application in various domains such as healthcare, finance, and transportation, for tasks like forecasts, anomaly detection, and classification, it is also important to discuss the interpretability of LLMs themselves and to improve their ability to discern cause-and-effect relationships \cite{jin2023large}.

In summary, LLMs are predominantly employed to produce explanations akin to traditional formats, enhanced with natural language integration. LLMs not only revolutionize the method of generating explanations but also how these processes and results are integrated into systems. More comprehensive strategies towards exploiting explainability in the LLM Era can refer to \cite{wu2024usable}. Further exploration is suggested into integrating LLMs as agents and connects LLMs to private data sources and APIs with frameworks such as LangChain and MetaGPT \cite{hong2023metagpt}.

\section{Human-Centered Explainable Interface in HAT}\label{sec:human}
 HAT has been defined as “the dynamic, interdependent coupling between one or more human operators and one or more automated systems requiring collaboration and coordination to achieve successful task completion” \cite{cuevas2007augmenting}. Teamwork among human involves three dimensions: cognition (team's mission and objectives understanding), skills (adaptability, communication, leadership), and attitudes (feelings about the task, mutual trust, and team cohesion) \cite{cannon1995defining}. Successful HAT requires mutual understanding of team members' attitudes, abilities, intentions, and states, along with a commitment to task completion and support \cite{chen2014human}. Additionally, cultural, experiential, and individual differences influence HAT performance \cite{schelble2020meaningfully}. Astrid Bertrand et al. \cite{Bertrand_2022} also discuss the impact of both normal and problematic cognitive biases on XAI design and evaluation, stressing the need for adaptive interfaces and the mitigation of biases.

\begin{figure}[htbp]
  \centering
  \includegraphics[width=\linewidth]{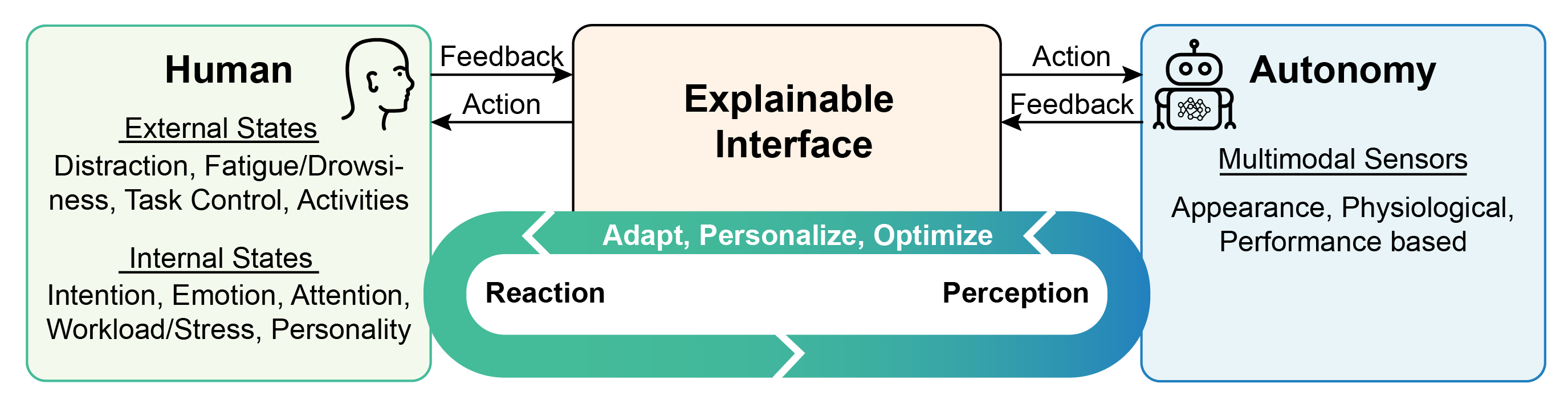}
  \caption{Interactive Dynamics between Humans and Autonomous Systems through an Explainable Interface.}
  \label{fig:interaction}
  \Description{This diagram presents the interaction loop between a human and an autonomous system, facilitated by an explainable interface. It's divided into three segments: The "Human" segment emphasizes both "External States" such as distraction, fatigue, and task control, and "Internal States" like intention, emotion, and stress, which affect the human's actions and feedback. The "Explainable Interface" segment acts as the mediator, designed to adapt, personalize, and optimize the interaction. It processes human feedback and actions and translates them into reactions and perceptions that the autonomy can understand. The "Autonomy" segment is equipped with multimodal sensors that capture various data points including appearance, physiological states, and performance metrics. The system uses this data to perform actions and provide feedback to the human. The continuous flow among these segments underscores the importance of adaptive feedback loops for creating a seamless human-autonomy interaction.}
\end{figure}

In Figure~\ref{fig:interaction}, the diagram depicts human-autonomy interaction mediated by an EI. The interface acts as a mediator, facilitating two-way feedback and action between humans and autonomy. The autonomy also perceives the human's reaction states, influencing the adaptability, personalization, and optimization of the EI. This continuous feedback loop is specifically designed to modeling improve team performance and collaboration.

\subsection{Human Behaviors Modeling in HAT}\label{sec:human-behaviors-modeling}
To build a complete human behavior understanding model for an EI in HAT, a variety of human states should be obtained, modelled, and incorporated. The current research for human modeling has been widely used in intelligent vehicles, autonomous piloting, and air traffic management \cite{xing2020advanced}. On the human side, it lists external states such as distraction, fatigue/drowsiness, task control, and activities, along with internal states like intention, emotion, attention, workload/stress, and personality. The autonomy side is characterized by multimodal sensors that may include data on appearance, physiological states, and performance-based metrics \cite{hu2022review}. Appearance-based sensors include the RGB camera, depth sensor, infrared sensor, etc. They are utilized to obtain facial and bodily movements. The physiological-based sensors monitor the various physiological factors, including the heart rate, brain waves, neuromuscular electromyography signals \cite{xing2024driver} and blood pressure. The performance-based sensors collect autonomy and task-related data during the operating process, such as the control and interaction quality, etc. Existing studies focus on a specific task and ignore the collaboration between various functions, thereby limiting recognition performance. Therefore, a unified human behavior model that considers both external and internal conditions is indispensable to endow the system with robust, precise, and time-variant sensing capabilities to assist the understanding of human behaviors, and hence contribute to better explanation qualities. 

Human distraction, often a result of secondary activities, is a key cause of errors and misinterpretations in team settings. Research in this area primarily focuses on using appearance-based detection and physiological data such as electrodermal activity, heart rate, breathing rate, and electroencephalogram (EEG) for early identification \cite{eraqi2019driver,kashevnik2021driver}. In critical and emergency situations, it is essential to communicate important signals clearly to heavily distracted operators. Incorporating audio or directional tactile warning signals can facilitate operators'rapid responses to potential collision events \cite{ho2007multisensorya,meng2015tactile}. 

Fatigue and drowsiness are another indication of human physical behaviors. Fatigue can be caused by either long-term repeated physical tasks or long-term monitoring and interpretation of autonomy's decisions and actions, which cause a heavy mental workload and mental fatigue \cite{diaz-garcia2021mental}. Detect human fatigue and drowsiness with eye-tracking technologies to monitor the variation and dynamics of the pupil is a common solution \cite{li2019proactive}. Also, EEG and electromyography-based data can be used to analyze the brain and muscle fatigues levels \cite{vollestad1997measurement}. To keep the human engaged and motivated, heavy fatigue must be avoided for the HAT, as heavy fatigue usually can be viewed as a key index of the failure of HAT \cite{rajavenkatanarayanan2019robotbased}. Thus, a key design requirement for EI is explaining the machine's thought by adapting to human's mental workload \cite{warren2020frienda}.

Regarding the impact of human psychological behaviors on HAT and EI designs, human intention is one of the critical aspects. Intention recognition is the process of understanding whether the ongoing activities of the agent are goal-oriented or not, and what is the goal behind these specific actions. Michon \cite{michon1985criticala} stated that the cognitive structure of human user-level intention can be divided into three sub-levels: strategical, tactical, and operational level. Understanding such an intention hierarchical plays an essential role in EI design for safer and more seamless HAT \cite{bayazit2003should}. Intention communication between humans and autonomy can helps building human's trust in the autonomy and identifying potential risks in advance. Similarly, understanding human intention can help autonomy better support the human and contribute to a higher team performance \cite{panganiban2020transparency,walliser2019team,tahboub2006intelligent}. 

Human attention estimation is another critical aspects for EI to support conditional teamwork and help the machine provide intervention or assistance in critical scenarios \cite{hu2022datadriven,jha2023estimation}. A typical approach for attention estimation is to build a mapping model between the context of the external scenario and the human field of attention based on deep learning \cite{ribeiro2019driver,ghosh2021speak2label}. When designing the human-attention-oriented EI, three basic benefits can be identified. First, estimate human attention to provide corrected cues to explanation contents. For example, if the human operator is highly focused on the task, abstracted key information can be provided to avoid a heavy workload. Second, it provides a cognitive correction and self-adjustment of autonomy. In this case, the autonomy can identify if the explanation satisfies the human's focus points and assess if the most expected explanation is provided. Last, estimating human attention enables the machine to learn human behaviors and the way human's percept current situations so that a human-like situation awareness process for autonomy can be learned.

Last, the emotional states of the human for HAT has been investigated in several studies \cite{monaco2020radarodo,bota2019review}. These studies indicate that the human emotional state critically impacts team performance and that the negative state is associated with aggressive collaboration that leads to poor team performance and even serious accidents. Adverse emotional states like frustration also prevent humans from receiving a correct understanding of autonomy, thereby reducing their sense of mastery, control, and task performance \cite{weidemann2021role}. There are two common types of approaches can be adopted to analyze human emotional states: discrete and continuous \cite{lerner2015emotion,gunes2011emotion}. The discrete approach categorizes emotional states into specific states such as happiness, surprise, sadness, and anger. The continuous approach uses a continuous space to indicate the emotional state. Emotion recognition has been studied from various perspectives with a variety of sensors using ML technology \cite{money2009analysinga}.

\subsection{Explainable Interface Design Principles}
\subsubsection{Considerations in Explainable Interface Design}
Survey \cite{methnani2024who} underscores the potential consequences of inadequate adjustments in the system's level of autonomy, such as errors, control conflicts, user frustration, and ultimately, the disuse of the system . Based on our previous discussion about human behavior, to ensure that EI-HAT can intuitively respond to human behavior and meet functional needs, it is essential to consider human-centered design principles by following approaches:

\textbf{Presets, Context and Situation Awareness for Common Ground:}
In XAI, the concepts of presets, context, and situation awareness are pivotal for establishing common ground between human operators and AI systems. The IMPACT program illustrates this through its use of high-level 'plays' as presets, simplifying complex multi-UxV control into manageable commands, thereby enhancing situational comprehension \cite{behymer2017initial}. In Transparent Value Alignment (TVA), the focus shifts to robots discerning human context and values, highlighting the significance of perceiving, and comprehending human actions to align robot behavior effectively \cite{sanneman2023transparent}. Meanwhile, the SAFE-AI framework delves deeper into situation awareness by categorizing it into perception, comprehension, and projection levels, each integral for AI systems to understand and explain their operational context to human users \cite{sanneman2022situation}. Collectively, these approaches underscore the importance of AI systems not only perceiving their environment but also interpreting and communicating this understanding back to human collaborators, thus establishing a shared common ground essential for effective and harmonious human-AI interaction.

\textbf{Personality and Adaptivity:}
Considering the vast cognitive diversity among human users, influenced by cultural contexts, personality traits, and past experiences, a universal approach to interface design may not suffice \cite{meske2022design,purificato2022first}. In a study investigating the effects of adaptive versus nonadaptive robots on information exchange and social relations, it was found that when the robot's dialogue was adapted to expert knowledge, expert participants perceived the robot as more effective, authoritative, and less patronizing \cite{torrey2006effects}. Furthermore, to ensure safe and efficient HAT, it's important to adjust trust in AI agents appropriately. Over-reliance on autonomous systems can lead to serious safety issues. Developing adaptive trust calibration methods is essential. These methods monitor users' reliance behavior and use cognitive cues, called "trust calibration clues," to detect and prompt recalibration of trust \cite{okamura2020adaptive}. Personalized interfaces cater to individual preferences and behaviors, rendering interactions more intuitive and relatable. For example, in the healthcare domain, clinical doctors and patients have significantly different needs for explanations, requiring different designs for EIs \cite{kim2024stakeholder}. Different patients have different preferences regarding the emphasis on negative emotions in explanations and their preference for detailed or holistic explanations. Exploring interactive features such as click or hover interactions or real-time personalized explanations through eye-tracking data can be beneficial  to increase system understanding and acceptance \cite{szymanski2022explaining}. 

\textbf{Interactivity:}
Interactivity is most mentioned among these considerations of EI design. It allows users to question, explore, and comprehend AI decisions, which promotes trust and alignment with human values. User feedback, visualization, and interactive tools are effective methods for enhancing transparency and facilitating user understanding of AI processes. For example, the Interactive Attention Alignment (IAA) framework improves user trust and understanding by fine-tuning DNNs to synchronize their behavior with human cognitive models, optimizing both prediction accuracy and attention mechanisms. \cite{gao2022aligning}. Additionally, methods such as incorporating user feedback for continuous AI learning, visualizing AI processes, and developing interactive tools that allow for the manipulation and exploration of AI models' decisions are important components. The introduction of Argumentative Exchanges (AXs) as a dynamic framework for interactive conflict resolution in XAI demonstrates the effectiveness of these methods \cite{rago2023interactive}. Furthermore, the integration of domain knowledge into AI development seeks to enhance both the design of EI and the accuracy of AI algorithms by implementing dialogical guidelines and knowledge acquisition \cite{gerdes2021dialogical}. Moreover, research in Interactive Machine Learning (IML) consolidates existing knowledge on interface design. The proposed structural and behavioral model for IML systems, along with the identified principles for effective interface building, emphasizes the ongoing necessity for research in this area, further underscoring the vital role of interactivity in creating AI systems that are transparent, understandable, and aligned with human-centric values \cite{dudley2018review,Teso_2023}.

\textbf{Multi-modality:}
The presentation and interaction of information in multiple modalities are also important for EIs. Currently, explanations are primarily conveyed through visual and auditory means, such as charts, heatmap, knowledge graphs, structured text, natural language, sound cues, and speech \cite{alipour2020study,lee2023investigating}. Although touch, smell, and even taste are important components of the human sensory system, they have received less exploration compared to the aforementioned modalities \cite{xu2023xair}.

Sharon Oviatt et al. \cite{oviatt2015paradigm} have discussed the transition from traditional Graphical User Interfaces (GUIs) to multimodal User Interfaces (MUIs), highlighting the advantages of MUIs in natural human-computer interaction and emphasizing their alignment with natural human communication processes. Dong Huk Park et al. introduced a model designed for simultaneous textual rationale generation and attention visualization based on two multimodal explanation modules, the Visual Question Answering Explanation (VQA-X) and the Activity Explanation (ACT-X) \cite{park2018multimodal}. MIRIAM, a prototype multimodal interface, provides voice assistance for operators. This system integrates natural language understanding, aiding in the development of clear and accurate mental models for autonomous systems. It demonstrates the potential of multimodal interfaces in improving understanding and trust for remote autonomous system operators in hazardous environments \cite{hastie2018miriam}. A study by Schneider et al. examines how different feedback modalities can improve the user experience in autonomous driving by testing feedback modalities like light, sound, visualization, text, and vibration. The results show that light and visualization feedback significantly enhanced the user experience in both practical and hedonic qualities compared to no feedback \cite{schneider2021increasing}. Holzinger et al. propose a holistic approach to automated medical decision-making that integrates ML research with human expertise. The authors introduce counterfactual graphs and emphasize the importance of incorporating multiple modalities throughout the decision pipeline by aligning local geometries across different input feature spaces \cite{holzinger2021multimodal}. 

Last, research on information olfaction plays a key role in human responses, despite not being a primary information channel \cite{patnaik2019information}. The study reveals that olfactory channels, including aspects like direction, saturation, airflow, and air quality, effectively augment information visualization. These channels are especially beneficial in enhancing memory, emotion, and immersion, and in providing clear signals when visual information is unclear or distracted. Additionally, olfaction complements other senses by delivering crucial information, such as specific scents for gas leak warnings or driver safety alerts \cite{grace2001drowsy}. Although olfactory cues are limited in providing complex explanations, they can be a potential vital element for more intuitive explanations for future multimodal EIs.

In conclusion, effective design principles for EIs hinge on four core considerations: establishing common ground through context awareness, personalizing adaptivity to accommodate user diversity, fostering interactivity for deeper understanding and trust, and integrating multi-modality for richer, more intuitive interactions. These principles underscore the necessity for EI designs that are not only technically advanced but also deeply attuned to human cognitive and sensory processes. By embracing these guidelines, we can develop HAT systems that are more accessible, transparent, and aligned with human needs, ultimately leading to more effective and harmonious human-AI collaborations.

\subsubsection{Extended User Experience Framework for Explainable Interface}
Susanne Bødker emphasizes the role of HCI from the perspective of human activity theory in 1989 \cite{bodker1989human}, this approach serves as the foundation of user centred design in HCI, advocating for interfaces that are intuitive, accessible, and tailored to the specific needs of users. For example, in the field of medical AI, the incorporation of Human-centered design approaches holds particular significance as it ensures that AI not only assists medical professionals in decision-making but also aligns with their workflow and cognitive processes \cite{holzinger2021human,xie2019outlining}.

The development of EIs has led to the formulation of guidelines and frameworks that bridge the gap between algorithm-centric and user-centred explanations \cite{liao2020questioning,weitz2021let}. These guidelines emphasis on involving UX and design practitioners in the creation of AI products that not only possess technical proficiency but also cater to user needs and contexts. Additionally, there is a need for tools that make it easier for designers and developers to design, develop, and evaluate EIs. For instance, ProtoAI introduces a new workflow called Model-Informed Prototyping (MIP) for designing AI-enhanced user interfaces, enabling designers to incorporate real-time AI model outputs into UI prototypes. This approach simplifies data-driven design, aligns with human-AI interaction principles, enables detailed prototyping with AI model insights, and supports interface simulation with various data and model scenarios \cite{subramonyam2021protoai}.

\begin{figure}[htbp]
  \centering
  \includegraphics[width=\linewidth]{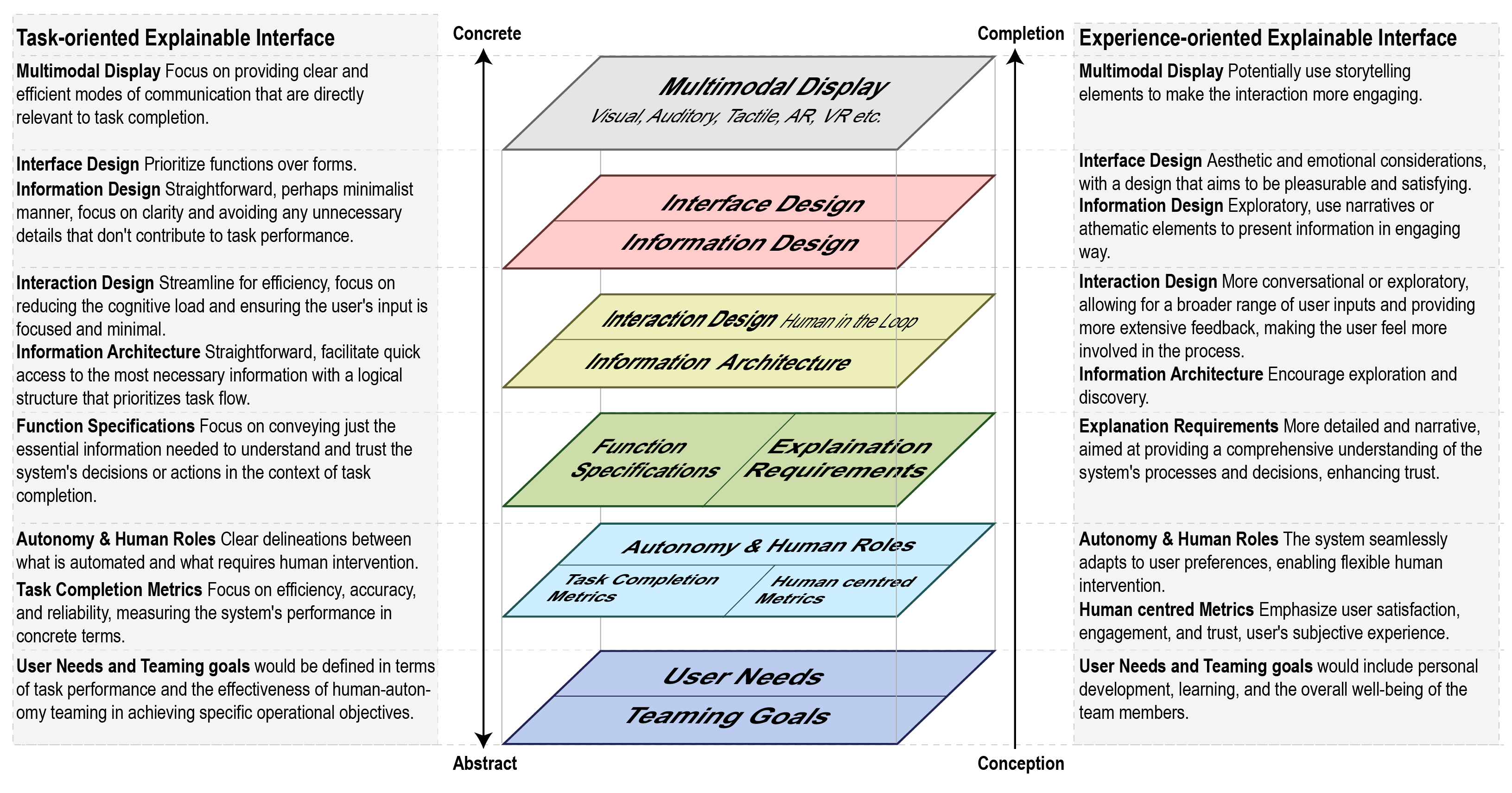}
  \caption{Extended User Experience Framework for Explainable Interface: Contrasting Task-oriented and Experience-oriented Approaches in Human-Autonomy Teaming.}
  \label{fig:UX}
  \Description{The figure depicts a framework comparing "Task-oriented" and "Experience-oriented" explainable interfaces in human-autonomy interactions. The task-oriented side focuses on efficiency and clarity for task completion, with streamlined design and interaction emphasizing functionality. The experience-oriented side aims to enrich user experience through engaging design, storytelling elements, and interactive exploration, prioritizing user satisfaction and well-being alongside task performance. Central to both is the adaptability to user needs and the integration of clear function specifications and autonomy roles.}
\end{figure}

Garrett introduced "The Elements of User Experience." The diagram delineates various levels of considerations that are essential for crafting a high-quality user experience (UX) \cite{garrett2011elements}. In the context of HAT, where automation takes different forms and operations are influenced by both human input and AI. As shown in Figure~\ref{fig:UX}, we propose the following extensions to this framework based on our earlier discussions, aimed at enhancing EIs within HAT.

\begin{itemize}
  \item \textbf{Task-oriented explainable interface:} Ideal for environments where the primary objective is to efficiently and accurately complete tasks, such as in industrial, medical, or military settings. 
  \item \textbf{Experience-oriented explainable interface:} Most suitable for scenarios where user engagement, long-term trust, and an intuitive understanding, such as in educational, customer service, or domestic environments.
\end{itemize}

The task-oriented perspective emphasizes efficiency, clarity, and practicality in achieving specific tasks, with a focus on the concrete aspects of the interaction. The Experience-oriented perspective, however, focuses on the quality of the interaction, aiming to enrich the user's experience and foster long-term benefits such as learning and personal growth. Both perspectives are valuable, and the optimal approach may involve balancing the two to create an interface that is both efficient and engaging for human-autonomy collaboration.

\section{Evaluation of Explainable Interface in HAT}\label{sec:evaluation}
In the HAT context, evaluating EI extends beyond traditional XAI methods, necessitating a human-centered approach that aligns evaluation methods with various roles' goals \cite{itoh2014design}. A paradigm shifts in XAI evaluation, as suggested in \cite{rosenfeld2021better}, prioritizes metrics that measure the explanation's relevance to the XAI's objectives, such as the discrepancy between the explanation's logic and the agent's actual performance, the number of generated rules, the complexity of the explanation, and its stability. Bhatt et al. \cite{bhatt2020evaluating} suggest three criteria for ideal feature-based explanations: low sensitivity to input changes, high faithfulness in identifying relevant features, and simplicity, all of which can be automated without human involvement. DARPA's framework \cite{gunning2019darpa} introduces measures including user satisfaction, mental model accuracy, task performance, trust assessment, and correctability. Different experimental conditions are used for evaluation, ranging from no explanation to full explanations for each decision and action, along with a control condition using a non-explainable system. Research \cite{vilone2021notionsa} categorizes explanation evaluation methods into objective and human-centered evaluations. Objective evaluations apply automated approaches and metrics to assess explainability, while human-centered evaluations focus on aspects like explanation quality, user satisfaction, comprehension, and performance. A distinction is made between the plausibility and convincingness of explanations versus their correctness, highlighting the importance of not conflating these criteria \cite{Nauta_2023}. Enyan Dai et al. \cite{dai2023comprehensive} outline essential qualities for explanations, such as correctness, completeness, consistency, contrastivity, user-friendliness, and causality, and suggests quantitative metrics for evaluating Graph Neural Network (GNN) explanations, including accuracy, fidelity, sparsity, and stability.

\begin{figure}[htbp]
  \centering
  \includegraphics[width=\linewidth]{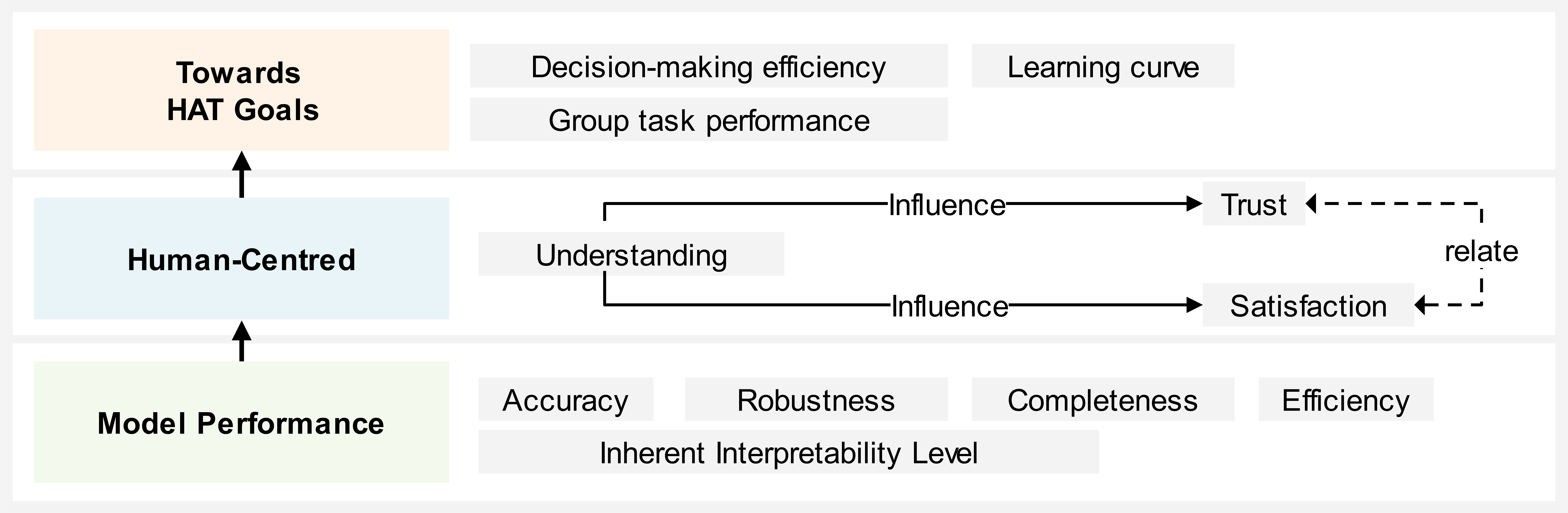}
  \caption{Evaluation Framework of Explainable Interface for Human-Autonomy Teaming.}
  \label{fig:evaluation}
  \Description{The figure outlines an evaluation framework for Human-Autonomy Teaming (HAT), linking model performance metrics like accuracy and interpretability to human-centered outcomes such as understanding, trust, and satisfaction. These elements collectively drive the achievement of HAT goals, including decision-making efficiency and group task performance, illustrating the interdependence of technical model performance and human factors in successful HAT systems.}
\end{figure}

Building on existing research, as shown in Figure~\ref{fig:evaluation}, we propose dividing evaluation metrics into three categories: model performance-centered, human-centered, and those aligned with HAT goals. This comprehensive evaluation framework is designed to ensure both technical efficiency and user-centric effectiveness in HAT systems, thereby promoting optimal integration and collaboration between humans and autonomous systems.

\subsection{Model Performance Centred Evaluation Methods}
\subsubsection{Inherent Interpretability Level Evaluation for Functional Model}
The inherent interpretability of a model largely hinges on its architecture. Simple models, such as linear regressions, small decision trees, or rule-based systems, are more interpretable due to their straightforward decision-making processes \cite{barredoarrieta2020explainable}. Within the same architecture, interpretability also varies with the model's complexity. In decision trees, fewer nodes or leaves indicate simpler decision rules, making the model more interpretable. Conversely, larger trees with more nodes capture complex patterns but are harder to understand and visualize. This complexity can be measured by the tree's number of nodes or depth \cite{souza2022decision}. Saliency or activation maps help interpret the focus of each layer in CNN or RNN \cite{crabbe2022concept,parvatharaju2021learning}. However, as the network depth increases, generating explanations becomes more challenging, especially for non-experts.In addition, when addressing complex tasks like vehicle state recognition and fault diagnosis, researchers often propose frameworks that integrate multiple basic models \cite{xing2023learning} to enhance performance. However, quantitatively comparing the interpretability of different architectures within such frameworks remains a more challenging endeavor.

\subsubsection{Quantitative Evaluation for Post-hoc Explainers}
\textbf{Accuracy} involves correctly representing the reasons for generating a specific output, particularly when the ground-truth rationale is known. It is closely linked to fidelity, which measures the faithfulness of identified substructures using heuristic metrics \cite{dai2023comprehensive}. CLEVR-XAI \cite{arras2022clevrxai} is a sizable visual question answering dataset featuring questions paired with pixel-level ground truth masks, serving as a benchmark for assessing visual explanations. In practical applications, the ground-truth, such as real feature importance, is often unknown. Simulating models and treating the system's output as ground truth is a common approach to assess explanation fidelity. This method utilizes ML metrics, including accuracy, precision, recall, F1-score, AUC, and others, to evaluate post-hoc explanations \cite{Mohseni_2021,ribeiro2016why}. Researchers also evaluated explanation accuracy by removing higher-order features and introducing noise to the inputs \cite{alvarezmelis2018robust}. Hugues et al. \cite{turbe2023evaluation} created a synthetic dataset with adjustable complexity and known discriminative features. They evaluated six popular explainers on both synthetic and real datasets like FordA and ECG. They introduced two new metrics, AUC$\widetilde{S}$top and F1$\widetilde{S}$, for assessing interpretability in time-series classification. AUC$\widetilde{S}$top measures the area under the curve as the top-k time steps are corrupted, reflecting explainers' accuracy in identifying important time steps.

\textbf{Robustness} is crucial for ensuring the stability and precision of AI explanations across varying conditions. Although a model's predictions might be consistent, its explanations can vary, highlighting the importance of robust explanations for reliable decision-making \cite{lai2019many}. Section~\ref{sec:model} outlines the heterogeneity in XAI explanation methods, such as feature importance and heatmap, complicating the identification of the most reliable method without a standard ground truth. To tackle this, Huang et al. introduce two novel black-box evaluation techniques for assessing interpretation discrepancy and probabilistic robustness. These include using a genetic algorithm for worst-case scenario analysis and subset simulation for assessing overall system robustness \cite{huang2023safari}. However, current post-hoc explainers exhibit limitations in robustness, especially in identifying biases in classifiers for sensitive applications. Common methods like LIME and SHAP often fail to detect biases related to protected attributes such as race \cite{slack2020fooling}. Furthermore, the robustness of GNN post-hoc explainers against label noise is another critical evaluation aspect, underscoring the need for accurate explanations in noisy data \cite{zhong2023robustness}.

\textbf{Completeness} in post-hoc explanations involves a balance between detailed insights into a model's internal mechanisms (reasoning-completeness) and the clarity and relevance of its outputs (output-completeness). Reasoning-completeness focuses on the depth of understanding a model's internal workings, where white-box models are essential for complete transparency in system operations. In contrast, output-completeness concerns how comprehensively the explanation represents the model's output \cite{Nauta_2023}. To assess output-completeness, the F1$\widetilde{S}$ score is used as a metric, providing a balanced evaluation of an explainer's ability to identify important time steps \cite{turbe2023evaluation}. However, while aiming for high completeness in explanations, it's crucial to strike a balance between sparsity and completeness for user comprehension. Explanations that are too sparse may omit critical information, whereas overly detailed ones can lead to complexity or redundancy, impeding user understanding.

\textbf{Efficiency} is particularly important in real-time or interactive systems, where prompt explanations are essential for maintaining user engagement and trust. In critical domains like medical diagnostics, financial decision-making, and autonomous vehicles, the ability to provide fast and clear explanations is crucial for timely decision-making, system trust, and avoiding potential harm or economic loss. Efficiency metrics include the speed of generating explanations and the computational resources needed. A proposed benchmark in \cite{idahl2021benchmarking} aims to address limitations in current explainers and foster the development of more effective, utility-driven methods. Study \cite{sun2023improving} indicates that gradient-based methods like GradCAM outperform perturbation-based methods in efficiency in medical image analysis. This efficiency stems from gradient-based methods requiring only a single input pass for explanation, as opposed to multiple perturbed inputs used by perturbation-based methods, leading to quicker processing times.

Above all, the model performance evaluation discussed can function independently of end-users, and its performance significantly impact the generation of explanations, which is a fundamental step in developing EIs.

\subsection{Human-Centered Evaluation Methods}
Different individuals perceive EIs distinctively. In coronary heart disease interface evaluations, non-experts showed higher satisfaction with various explanations compared to experts \cite{panigutti2023codesign}. Mohammad et al. found that in clinical decision support systems, example-based and counterfactual explanations were more comprehensible and less likely to foster over-trust in AI than local and global ones \cite{ghai2023fair}. The study \cite{schraagen2020trusting} examines the effects of self-driving car explanations on user trust and understanding. It found that causal explanations were least satisfying, intentional explanations increased trust the most, and mixed explanations yielded the best overall understanding and consistent trust. These findings illustrate that EI-HAT differs from general model explainability, with no universal best explanation applicable across scenarios and audiences. These highlight the necessity for human-centered evaluation such as understanding, trust, and satisfaction.

\textbf{Human understanding} makes a basic of HAT in complex, dynamic environments because it is essential to build a shared mental model for human and autonomy. Situation Awareness (SA) playing a key role in this process \cite{mutzenich2021updating}. For a lower level of autonomy, it's important to maintain human engagement and responsibility for the task, with the autonomous system providing warnings and support via sophisticated assistance technologies \cite{on2021taxonomy}. Paul Salmon et al. \cite{salmon2006situation} propose that the achievement and maintenance of SA is influenced by both individual and task factors, such as experience, training, workload and also interface design. The method for measuring SA must be consistent and reliable, producing comparable assessments under similar circumstances. Evaluating SA should satisfy three core criteria: it should enable simultaneous measurement across different situations, assess awareness on both an individual and team level, and provide insights in real-time. 

\textbf{Human trust} is shaped by the user's AI knowledge, task understanding, and objectives. EIs are designed to foster appropriate trust, aligning with AI's capabilities to avoid both under utilization and excessive reliance. For instance, excessive trust in autonomous driving could lead to neglecting necessary manual interventions, risking accidents \cite{lee2004trust}. In autonomous vehicles, establishing trust requires understanding the system's abilities and limitations \cite{haspiel2018explanations}. Trust is sensitive to system errors and tends to fluctuate over time \cite{basso2016trust}. Measuring human trust in EIs can be a key indicator of their effectiveness, looking at how trust is built and adjusted through various types of explanations \cite{hoffman2018metrics}. Trust is connected to team interaction and failure recovery, suggesting that it is a dynamic process best measured through real-time, sensor-based, and behavior-based methods, beyond traditional surveys \cite{demir2021exploration}.

\textbf{Human satisfaction} with an EI doesn't necessarily imply correct system understanding or trust \cite{hoffman2023measures}. Satisfaction with explanations is determined by the extent to which users perceive the ease of use and utility of an AI system or process. Key characteristics of satisfactory explanations include completeness, actionability/persuasiveness, usefulness, relevance, personalization, and timeliness. These attributes can be measured using various methods like surveys, questionnaires, user interviews, feedback after interaction, user experience metrics, metrics from recommendation systems, and assessments of contextual relevance \cite{tekkesinoglu2024exploring}. Ashraf Abdul et al. \cite{abdul2020cogam} introduced the Cognitive-GAM (COGAM) framework, which formalizes cognitive load for XAI by using visual chunks and aims for a balance between cognitive load and accuracy in generating explanations.

Measuring these aspects effectively in practice necessitate active participation from individuals. Comparative experiments are key in evaluating EIs, by contrasting interfaces with and without explanations or examining different types of EIs. Hoffman's research \cite{hoffman2023measures} delves into the explanation process, highlighting curiosity's role, and introduces a detailed set of scales to measure aspects like explanation quality, satisfaction, mental models, and trust, useful in question-and-answer assessments. Myounghoon Jeo's research \cite{jeon2023effects} suggests that AI systems evoking positive emotions can boost trust. Mishra et al. \cite{mishra2018cognitioncognizant} developed a hierarchical LSTM model using eye-tracking data for sentiment analysis in text. Mengyao Li et al. \cite{li2022estimating} proposed using conversational data, focusing on lexical and acoustic features, to assess trust dynamically in NASA missions. Advanced affective computing techniques involve emotion recognition through text, audio, and visual cues, as well as physiological data like EEG or ECG, often analyzed using ML \cite{wang2022systematic}. The DEVCOM Army Research Laboratory's HAT3 toolkit \cite{neubauer2023human} integrates various indicators, including behavioural and physiological cues, for trust analysis in human-autonomy interactions. 

While human-centered evaluation methods are vital, focusing exclusively on one aspect, like high trust levels, is not always indicative of superior results. It's important to value both objective and subjective measurements. Affective computing frequently utilizes emotional models, including Ekman's six basic emotions \cite{ekman1971universals}, Plutchik's wheel \cite{plutchik2003emotions}, or the Pleasure-Arousal-Dominance (PAD) model \cite{bakker2014pleasure}, and is supported by well-developed multimodal databases \cite{wang2022systematic}. However, the relationship between these emotions and aspects like understanding, trust, and satisfaction need further exploration. What's more, in HAT, Dynamic evaluation of human-centered metrics such as trust calibration can allow for continuous improvement and adaptation of the system to better support the team's needs.

\subsection{Evaluation Methods towards Teaming Goals}
Human-centered evaluation are not always align with the performance of HAT \cite{hoffman2023measures}. Though they are important, the ultimate goal is to successfully complete tasks. Systems that fail to enhance collective goal achievement, despite advantages in interpretability and satisfaction, are deemed suboptimal. This underscores the importance of evaluating EIs in the context of HAT objectives. Factors such as learning curves, decision-making efficiency, and group task performance play a critical role in the dynamics of HAT.

The learning curve can be evaluated by measuring the performance of the HAT on specific tasks. Over time, the team should become more proficient, completing tasks more quickly and with fewer errors \cite{ferrier2022measuring}. The learning curve can be evaluated through various measures such as task completion time, error rates, physiological indicators, psychological factors (such as self-reported affective state and stress), and perceived workload and team efficiency. Characteristics of autonomous agents, team composition, task attributes, human individual difference variables and training significantly influence the HAT performance \cite{oneill2022humana}.

Group task performance represents the most direct measure of their effectiveness or usability. The Perceptual-Cognitive Explanation (PeCoX) framework, which aims to improve HAT performance, underscores the importance of generating, conveying, and receiving explanations \cite{neerincx2018using}. When considering EIs within this framework, they act as a bridge facilitating team interaction. Thus, assessing EIs must focus on key aspects of team performance, including the efficiency, completeness, and quality of task execution. These factors offer tangible metrics to evaluate how EIs enhance the overall effectiveness and collaborative success of the HAT system.

\section{Challenges and Future Directions}\label{sec:challenges}
\subsection{Challenges}
Incorporating XAI into HAT systems and facilitating interaction through EIs poses a variety of critical challenges:

\begin{itemize}
  \item \textbf{Timely explanation generation:} In high-stakes, real-time HAT settings, it's essential for EIs to offer timely and understandable explanations in proper time to foster appropriate trust and assist in decision-making.
  \item \textbf{Explanation generation methods selection:} The right method for generating explanations is complex and depends on factors like data complexity, target audience, and context of use. The range of methods varies from simple rule-based systems to complex ML models, and the use of different post-hoc explainers can lead to inconsistent results \cite{krishna2022disagreement}.
  \item \textbf{Explanation presentation methods selection:} How explanations are presented in EIs is as important as the explanations themselves. The presentation method can greatly affect HAT performance. The same information can be perceived differently depending on whether it is presented via text, visuals, audio, tactile, or olfactory means. The choice of presentation method varies with the task type and environment, requiring careful selection to ensure effectiveness and appropriateness for the specific context and user needs.
  \item \textbf{Complexity in designing explainable interface frameworks:} Creating these frameworks is an interdisciplinary effort, requiring knowledge from AI, HCI, psychology, and relevant domains. This necessitates a collaborative approach that extends beyond traditional disciplinary limits, Personalizing explanations to accommodate individual differences is essential \cite{wintersberger2020explainable}.
  \item \textbf{Personalized and adaptive design:} Interfaces should be tailored to users' varying cognitive abilities and expertise levels. This involves modeling human behavior dynamics and adjust EIs accordingly, thereby fostering accurate understanding, reasonable trust, and increased satisfaction.
  \item \textbf{Hybrid dynamic evaluation:} As discussed in Section~\ref{sec:evaluation}, evaluating these interfaces requires a balanced approach. This includes assessing the model's performance, the human-centered evaluation, and the overall performance of the HAT system. Methods like questionnaires and affective computing should be utilized, and more empiral research is needed for specific tasks and domains.
  \item \textbf{Preparing users for HAT through explainable interfaces:} It is vital to equip users with the necessary knowledge about the system for effective teaming. The design of EIs that can achieve this educational goal is an area that requires further exploration.
  \item \textbf{Multi-disciplinary collaboration:} Incorporating EIs into HAT systems goes beyond technical considerations. It requires viewing autonomy as a participant in human society and designing responsible AI ecosystems. Bridging the gap between responsible AI research, industry, policy, and regulatory efforts necessitates a comprehensive integration of art, humanities, and other disciplines \cite{braid}.
\end{itemize}

\subsection{Directions}
To effectively enhance HAT systems through EIs, a comprehensive and structured human participated approach is essential. The approach should commence with a re-evaluation of traditional prototype design and software development processes, prioritizing the seamless integration of XAI either as a distinct module or within the functional interface. A pivotal element of this strategy involves establishing a decision framework. This framework should facilitate the selection of explanation methods that align with the intricacies of the data, the user's comprehension level, and the contextual application of the system.

The debate between the use of interpretable models and post-hoc explanations for black-box models has gained attention in recent research \cite{rudin2019stopa}. Efforts have been made to explore neuro-symbolic AI, which aims to integrate the continuous and discrete aspects of neuro and symbolic AI \cite{belle2024neurosymbolica}. Additionally, causal inference and IML approaches have been examined \cite{luo2020whena, Teso_2023}. However, despite the emergence of more interpretable model architectures, there are challenges in designing systems that effectively explain these models to humans. Moreover, it is crucial to focus on developing user-friendly tools for interactive post-hoc explainer selection, allowing non-AI experts to participate in system design \cite{han2022explanation}.As discussed in section \ref{sec:llm}, the utilization of advanced multimodal LLMs
to explain simpler black box models is another valuable direction, providing human-comprehensible explanations and
simplifying the EI development process. However, the application of LLMs in design, the mitigation of potential
issues arising from LLM hallucinations to produce reliable explanations, and the interpretability LLM itself remain areas for exploration.

For explanation presentation, it is good to draw on insights from visualization and conduct empirical researches. These researches should evaluate the impact of multimodal content presentation methods, like visual, textual, or auditory formats, on user understanding and interaction. When establishing bridges between autonomy and humans through EIs, considering presets, along with incorporating interactivity, adaptability, and multimodality, has the potential to improve the performance of HAT. To ensure comprehensive evaluation, a hybrid dynamic multimodal approach is recommended. This combines quantitative methods like model performance metrics with qualitative methods such as user feedback and surveys. Additionally, integrating affective computing to assess human behavior provides valuable insights for interface refinement.

\section{Conclusion}\label{sec:conclusion}
Drawing on the survey findings, this paper emphasizes the necessity for ongoing research to improve aspects like understandability, trustworthiness, user satisfaction, and the overall efficiency and effectiveness of XAI-enhanced HAT systems. We clarify crucial concepts such as explainability, explanations, and EIs within HAT systems, help researchers first step into this field from various domains to discern the current state, challenges, and potential focal points. The paper serves as a valuable resource for stakeholders involved in XAI systems, proposing a design framework for EIs. It also introduces an elaborate evaluation framework that containing model performance, human-centric factors, and HAT task objectives. This framework deconstructs the intricacies involved in assessing EIs and XAI systems. Furthermore, we specifically detail the challenges and potential avenues for future research in this area both from the explanation generation and the design and integration of them within the system by EIs.

\bibliographystyle{ACM-Reference-Format}
\bibliography{sample-base}

\appendix
\title{Online Appendix to: Explainable Interface for Human-Autonomy Teaming: A Survey}
\section{EXPLAINABLE INTERFACE APPLICATION SCENARIOS IN HAT: CASE STUDIES}\label{sec:application}
Thougn EIs play a pivotal role in enhancing the efficiency of HAT in high-risk sectors like healthcare, transportation, and manufacturing, there is a lack of empirical studies examining end-users' explainability needs and behaviors around XAI explanations in real-world contexts \cite{kim2023help}. Research \cite{beer2014toward} outlines a framework to evaluate robot autonomy levels in human-robot interaction (HRI). It redefines autonomy as a robot's ability to perceive, plan, and act in its environment for specific tasks. A 10-point taxonomy categorizes levels of robot autonomy (LORA), assessing how these levels influence key HRI factors like acceptance and reliability. The study also underscores autonomy's interplay with constructs like safety and trust in HRI, essential for comprehending autonomy's role and impacts in these interactions. We summarize related researches as Table~\ref{tab:levels} which illustrates how, across varying levels of autonomy, the roles of humans and autonomous systems differ, thereby influencing the functionality of EIs in these contexts. Notably, the concept of variable autonomy in HAT for firefighting, introduced in \cite{verhagen2024meaningful}, allows for the dynamic adjustment of robot autonomy based on situational demands. This approach leads to efficient task sharing and role adaptation between humans and robots, implying a need for similarly dynamic EIs in these contexts.

\begin{table}
  \centering
  \caption{Design of Explainable Interface for Different Autonomy Levels}
  \label{tab:levels}
  \begin{tabularx}{\textwidth}{>{\centering\arraybackslash\hsize=0.5\hsize}X>{\raggedright\arraybackslash\hsize=1.17\hsize}X>{\raggedright\arraybackslash\hsize=1.17\hsize}X>{\raggedright\arraybackslash\hsize=1.17\hsize}X}
    \toprule
    \textbf{Level} & \textbf{Human Role} & \textbf{Autonomy Role} & \textbf{Functions of EIs} \\
    \midrule
    Low Autonomy & Direct Controller: Executes immediate control, frequent commands, continuous monitoring & Active Executor: Acts upon human input nearly real-time, no independent decision-making. & Clarify operational mechanisms and respond to direct human commands. \cite{beer2014toward,berlin2006taxonomic} \\
    \hline
    Medium Autonomy & Supervisor/Guide: Involves setting parameters, providing intermittent input, and intervening in complex decisions beyond the agent's programming. & Assistant: Offers insights into decision-making under specific conditions & Provide insights into decision-making, especially under specific conditions or parameters. \cite{beer2014toward,lieberman2020roboticassisted} \\
    \hline
    High Autonomy & Overseer/Monitor: Engages in high-level monitoring and intervenes in unexpected situations. & Self-regulated Actor/Adaptive Manager: Performs tasks independently, seeks input when uncertain & Offer detailed explanations of complex decisions, highlighting the influencing algorithms, behaviors, or data. Justify actions retrospectively \cite{lyons2018exploring}.  \\
    \hline
    Full Autonomy & Initializer/Auditor: Establishes initial parameters and high-level goal setting. Beyond the setup, humans are mostly hands-off, only intervening in exceptional cases or for maintenance. & Sovereign Operator: Manages all tasks autonomously, constantly improving or adapting its strategy based on goals and feedback from its environment. Human input is rare. & Explain overall strategies and objectives, clarify adjustments and rationale in autonomous strategies, and ensure consistency between human intentions and autonomous actions. \cite{lyons2018exploring} \\
    \bottomrule
  \end{tabularx}
\end{table}

\subsection{Healthcare}
\subsubsection{Clinical Decision Support Systems}
EIs are vital in Clinical Decision Support Systems (CDSS) for clinician acceptance and transparency of AI'decision-making is more effective than merely presenting AI confidence levels \cite{panigutti2022understandinga}. Factors influencing the effectiveness of explainability in CDSS include technical integration, algorithm validation, deployment context, decision-making roles, and user demographics \cite{amann2022explain}. 
Computer-based CDSS differ in operation, knowledge sources, and workflow impact. A substantial portion focuses on patient-centered decision support, while a smaller percentage provides online support for clinicians. Most require additional personnel for data management \cite{berlin2006taxonomic}. Implementing CDSS involves considering delivery methods, governance, incentives, workflow integration, content updates, and data quality \cite{wright2010best}. AI-based CDSS are evolving with complex algorithms, analyzing large datasets for precise, personalized support, improving healthcare quality and efficiency. However, issues such as data privacy, transparency of algorithms, and integration into existing workflows present ongoing challenges \cite{ramgopal2023artificial}. Key elements in the design and implementation of EIs for CDSS include:
\begin{itemize}
  \item Real-time visualization: EIs should display system's decision-making processes and actions in real-time, allowing users to observe and understand how the system arrives at its conclusions. This feature enables clinicians, operators, or users to monitor and intervene when necessary \cite{amann2022explain}.
  \item Alerts and notifications: EIs generate alerts and notifications to inform users about potential complications, anomalies, or critical events. This feature enables timely intervention by support staff, who can take appropriate actions based on the provided information. 
  \item Context-aware assistance: EIs should deliver situation-specific information, like patient condition, current procedures, and relevant literature, aiding support staff, nurses, and anesthesiologists in recognizing potential issues and making informed decisions \cite{yang2023harnessing}. External Explanations also matters, such as biomedical literature or prediction process allow clinicians to validate system reasoning against their own knowledge and determine if recommendations are clinically appropriate \cite{pierce2022explainability,yang2023harnessing}.
  \item Natural language explanation: Providing natural language explanations and visualizations can help clinicians understand the robot's control logic and reasoning, improving trust and confidence in the technology \cite{Markus_2021}. Study \cite{umerenkov2023deciphering} shows that LLMs in CDSS can clarify medical diagnoses but must be used cautiously.
  \item User-Centered Design: Emphasizing the importance of aligning with clinical workflows and user needs, human-centered design is crucial in CDSS development \cite{wang2023humancentered}. This approach involves iterative processes, including prototyping, testing, and redesigning \cite{panigutti2023codesign}.
\end{itemize}

\subsubsection{Surgical Robotics}
The development of autonomy in surgical robots can be categorized based on a combination of human involvement and the robot's level of proficiency. At Level 0, the robot is completely under the control of the surgeon, serving merely as a tool. In contrast, Level 5 envisions a robot performing an entire surgery independently. Intermediate levels involve the robot transitioning from providing mechanical assistance to autonomously executing specified tasks, suggesting surgical methods awaiting human confirmation, and even making medical judgments under the supervision of a doctor \cite{yang2017medical}. This range from Level 0 to Level 5 encapsulates a progression from total human oversight to complete self-sufficiency of the robot. Currently, most surgical robots operate at the level 1 and level 2, with the surgeon playing a significant role in the procedure. However, advancements in AI and sensors are improving the precision, grip, feedback, and autonomy of robotic surgery systems \cite{fiorini2022concepts}. 

\begin{figure}
  \centering
  \includegraphics[width=\linewidth]{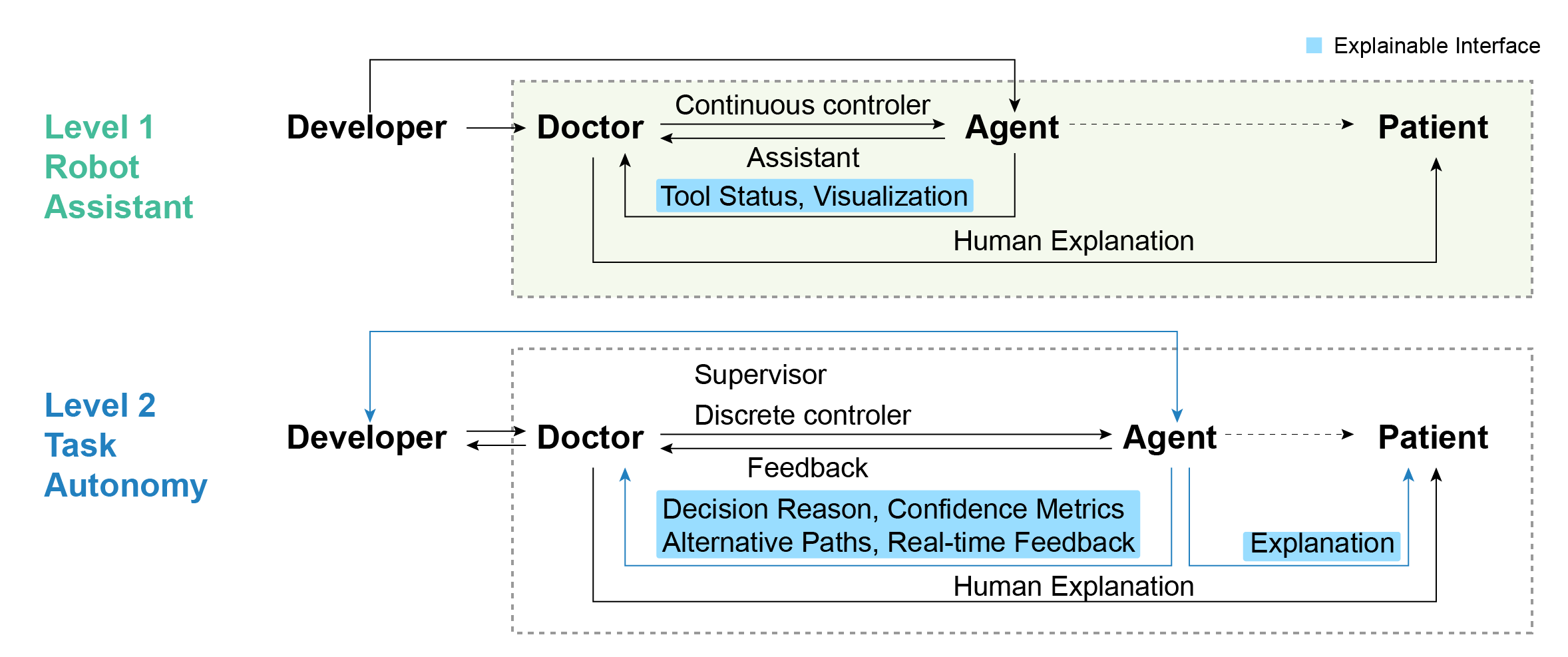}
  \caption{Explainable Interface in level 1 and level 2 surgical robotics}
  \label{fig:healthcare}
  \Description{Figure 1 illustrates Level 1, 'Robot Assistant', where a continuous interaction loop exists between
  the developer and the doctor. In this level, EI provides tool status updates and enhanced visualiza-
  tions, like 3D renderings of the surgical area, for clearer insights. Level 2, 'Task Autonomy',
  positions the doctor in a supervisory role. Here, control over the robot is discrete rather than
  continuous.}
\end{figure}

Figure~\ref{fig:healthcare} illustrates Level 1, 'Robot Assistant', where a continuous interaction loop exists between the developer and the doctor. In this level, EI provides tool status updates and enhanced visualizations, like 3D renderings of the surgical area, for clearer insights \cite{dappa2016cinematic}. Level 2, 'Task Autonomy', positions the doctor in a supervisory role. Here, control over the robot is discrete rather than continuous. An application of this is in spinal surgery, where robotic guidance is integrated with preoperative planning software \cite{lieberman2020roboticassisted}. EIs for doctor include feedback on decisions, reasons, confidence levels, alternative suggestions, and real-time updates. In the future, it is possible that surgical robots will achieve higher levels of autonomy, allowing them to perform more complex tasks independently while still being supervised by human surgeons. However, there also are researches arguing for a shift in focus from complete autonomy to enhancement of human capabilities in the context of autonomous surgery \cite{battaglia2021rethinking}. As AI becomes more integrated into healthcare, challenges arise, especially concerning explainability, utility, trust, legal, regulatory, and ethical \cite{yang2017medical,osullivan2019legal,wysocki2023assessing}. 

The surgical robotics field is characterized by varied interaction methods to enhance surgical precision or improve experience. Developments include robots capable of environment reconstruction and force feedback \cite{li2018surgical}.  Minimally invasive surgery primarily consists of a series of sub-tasks, which can be broken down into basic gestures or contexts. Surgical gesture recognition can aid in motion planning and decision-making, as well as establish context-aware knowledge to improve the control quality of surgical robots \cite{zhang2021surgical}. Furthermore, as robots become more autonomous, the importance of strategic task planning becomes increasingly clear \cite{ginesi2020autonomous}. There is a trend towards optimizing shared autonomy, advocating for a more collaborative approach between humans and robots. This human-in-the-loop collaborative method enhances human capabilities alongside robotic autonomy \cite{gopinath2017humanintheloop}. A study reviews current surgical robotic interfaces, highlighting advancements and limitations in feedback and visualization, and proposes a novel design featuring bedside access, sterile procedure, 3D head-mounted visualization, and intuitive control for enhanced surgical outcomes \cite{simorov2012review}. 

\subsubsection{Challenges fo Explainable Interface in Healthcare}
Based Advancements in healthcare suggest that their future depends not only on technological progression but also on creating interfaces that integrate seamlessly with AI, prioritizing clarity and comprehensibility. Design and implementing EIs in healthcare faces unique challenges:
\begin{itemize}
  \item Complexity of Medical Data: The inherently intricate and varied nature of medical data, including electronic health records, medical imaging, genomics, and sensor data, presents a significant challenge. XAI systems must be adept at interpreting and explaining decisions based on this diverse, high-dimensional data \cite{arora2023value}. Furthermore, the lack of standardized datasets exacerbates the difficulty of model training.
  \item Real-time Clinical Decision Making: Given the high stakes involved in healthcare decisions, the critical nature of surgical procedures, the necessity for real-time decision-making is paramount.\cite{amann2022explain}.
  \item Varied User Expertise: The range of technical and medical knowledge among XAI system users in healthcare, such as doctors, nurses, and patients, varies widely \cite{kim2024stakeholder}. Crafting explanations that are understandable to this diverse user base is a considerable challenge, necessitating customization to suit different comprehension levels.
  \item Regulatory Compliance and Ethical Considerations: Healthcare's strict regulatory environment demands compliance, especially as the FDA starts to view certain AI models as medical devices, necessitating stringent regulatory frameworks \cite{health2023artificial}.
  \item Integration with Clinical Workflows: The effectiveness of EIs in healthcare hinges on their seamless integration with existing clinical workflows. This necessitates interdisciplinary collaboration to ensure that the implementation of these interfaces does not introduce additional complexity or time-consuming burdens for healthcare professionals \cite{panigutti2023codesign}.
  \item Evaluating Explanation Effectiveness: The difficulty in evaluating healthcare explanations lies in their subjective and objective combined nature, as effectiveness may vary based on their role in clinical decision-making, patient understanding, and improving outcomes.
  \item Data Privacy and Security: Ensuring the privacy and security of sensitive health data used in XAI systems is paramount. This includes addressing concerns related to data sharing, consent, and anonymization \cite{albahri2023systematic}.
\end{itemize}

\subsection{Transportation}
\subsubsection{Autonomous Vehicles}
HAT in transportation and autonomous vehicles (AVs) can be categorized into six levels of driving automation, as defined by the Society of Automotive Engineers (SAE) \cite{on2021taxonomy}. As the level of automation increases, the role of the human driver shifts from active control to supervision. This transition can lead to challenges in maintaining driver engagement, situation awareness, and trust in the automation system \cite{beggiato2015would}. Level 3 and higher AVs are equipped to handle full control, allowing drivers to focus on non-driving activities. However, human intervention is still required in certain situations. Providing explanations within AVs enhances the driver's sense of control and understanding of the vehicle's status and capabilities, reducing potential anxiety. These intervention needs differ depending on non-driving tasks, driving conditions, and takeover methods. Personalized AVs, tailored to individual driver behaviors, enhance reliability and foster greater trust in the system \cite{salubre2021takeover,kyriakidis2019human}. Shahin et al.'s survey about XAI in autonomous driving, proposes an framework for a safer, transparent, publicly approved, and environmentally friendly intelligent vehicles \cite{atakishiyev2023explainable}. As self-driving cars become more prevalent, regulatory bodies might see the value in having these types of interfaces \cite{xu2023enabling}. Implementing such interfaces not only be beneficial for users but can also help manufacturers meet demands and gain approvals from regulatory agencies \cite{biondi2019human}. To present EIs inside and outside vehicles, visual, auditory, and haptic/tactile aspects should be considered:

\begin{itemize}
  \item Visual interface: Visual interfaces in AVs have evolved from traditional physical controls to advanced touchscreens and augmented reality head-up displays (AR-HUD). They enable passengers to set destinations, monitor progress, environment, vehicle status, and manage settings like climate and entertainment, thereby improving efficiency in control takeover, enhancing user satisfaction, and trust \cite{oliveira2018evaluating,jing2022impact}.
  \item Auditory interface: Speech output is highly relevant to improve driver attitudes that affect acceptance of automated systems is brought forth \cite{forster2017increasing}. In AVs, auditory user interfaces are designed to ensure safe control transitions, especially during the conditional self-driving phase. These interfaces use a combination of voice and sound feedback to communicate with the user. Compared to generic auditory output, communicating upcoming automated manoeuvres additionally by speech led to a decrease in self-reported visual workload and decreased monitoring of the visual HMI \cite{naujoks2017improving}. 
  \item Haptic/tactile interface: AVs can enhance user experience and control through the integration of haptic feedback interfaces. A study compared two types of driver-vehicle interfaces for controlling vehicles: one using a haptic interface with both kinesthetic and tactile feedback, and the other using a hand-gesture interface with augmented reality feedback. Compared to manual driving, semi-autonomous control significantly reduced the perceived workload. The haptic interface, in particular, proved to be especially reliable and effective in comparison to the gesture interface, as it provides direct and immediate feedback to the driver \cite{manawadu2016haptic}.
\end{itemize}

\begin{figure}
  \centering
  \includegraphics[width=\linewidth]{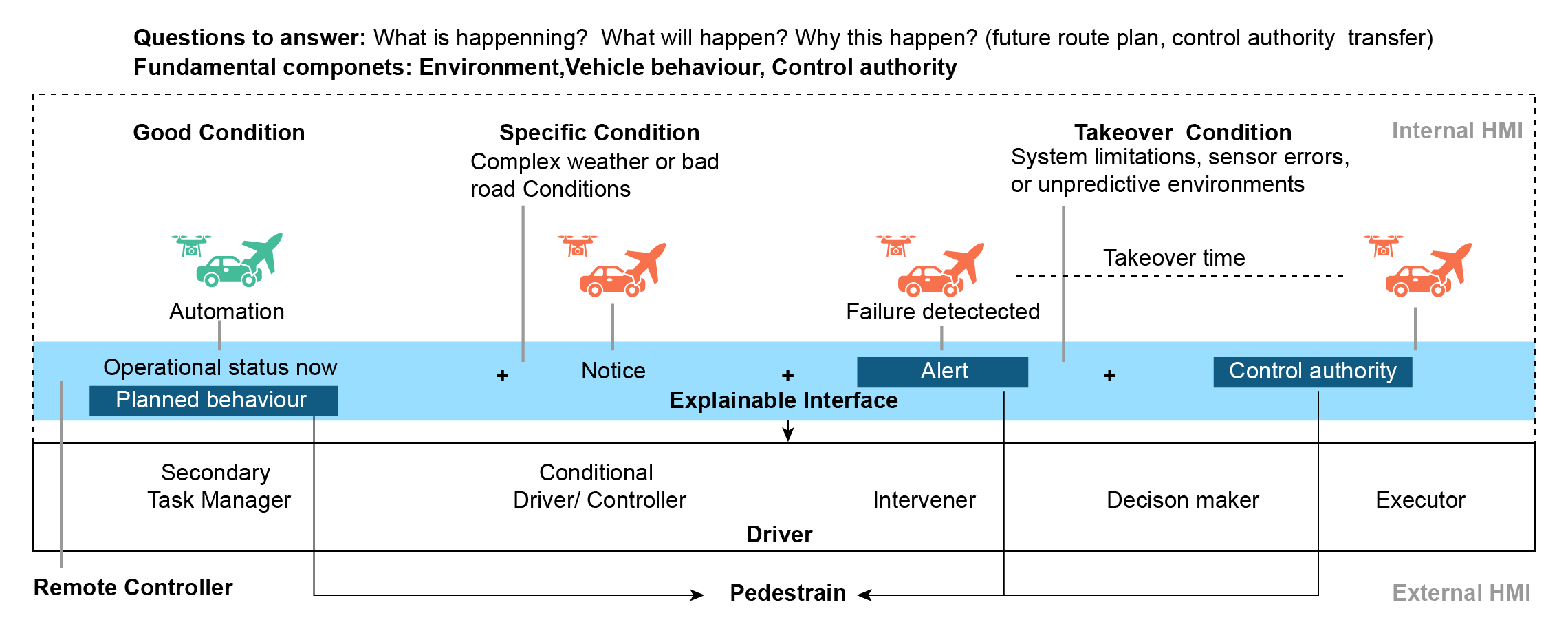}
  \caption{Explainable Interface Requirements for Human Vehicle Interaction in Conditional Driving Automation}
  \label{fig:AVs}
  \Description{Figure 2 delineates the requirements for EIs in AVs across three operational conditions: 'good condition' for standard autonomous operation, 'specific condition' for complex scenarios like adverse weather requiring increased notifications, and 'takeover condition' triggered by system errors or unpredictable circumstances necessitating control transfer. EIs should avoid overloading users with complex technical details, instead clearly presenting current and anticipated environmental conditions, vehicle behaviors, and control information. The interface should also allow users to retrospectively track significant environmental changes, behaviors, and control shifts during the journey.}
\end{figure}

Multimodal explanations utilizing two types of displays are more effective than unimodal or three-display multimodal explanations, with speech-based displays producing the best takeover performance. The combination of visual, auditory, and tactile cues, especially visual-auditory warnings, improves response efficiency in scenarios like vehicle malfunctions and highway exits \cite{lee2023investigating,yun2020multimodal}. The MIRIAM study demonstrates a multimodal interface combining visual indicators with a conversational agent, maintaining situation awareness across various cognitive styles and enhancing user engagement through natural language interaction \cite{robb2018keep}. The timing of information delivery is as crucial as its content, with traffic risk probability guiding when explanations are provided, particularly to avoid overwhelming passengers with excess information \cite{kim2023what}. Figure~\ref{fig:AVs} delineates the requirements for EIs in AVs across three operational conditions: 'good condition' for standard autonomous operation, 'specific condition' for complex scenarios like adverse weather requiring increased notifications, and 'takeover condition' triggered by system errors or unpredictable circumstances necessitating control transfer. EIs should avoid overloading users with complex technical details, instead clearly presenting current and anticipated environmental conditions, vehicle behaviors, and control information. The interface should also allow users to retrospectively track significant environmental changes, behaviors, and control shifts during the journey. Incorporating predictive maintenance data, such as remaining useful life and time-to-failure, is important \cite{shukla2020opportunitiesa}. As for remote control, more comprehensive dynamic team situation awareness is needed \cite{mutzenich2021updating}. 

In evaluating autonomous systems, it is imperative to assess understanding, trust, and satisfaction in real-time interactions during the operation of AVs. This approach ensures more accurate and useful results compared to traditional questionnaires, which may not be effective in this context. Measurements should integrate the use of existing vehicle devices, such as cameras, and wearable devices capable of monitoring human psychological activities for dynamic modelling of human behavior in HAT.

\subsubsection{Air Transportation Systems}
The integration of EIs is pivotal in enhancing the functionality and safety of transportation systems. The European Union Aviation Safety Agency (EASA) is actively working towards developing guidelines for AI applications intended for safety-related applications in aviation. The stakeholders, including air traffic controllers, airport operators, and Uncrewed Traffic Management (UTM) service providers, seek assurance that the AI solutions are trustworthy and safe \cite{torensguidelines}. Developing robust solutions to manage, measure, and maintain trust between humans and autonomous systems is important, particularly in maintaining existing safety standards in aviation.

In Air Traffic Management (ATM), XAI is instrumental in managing air traffic flow and capacity, ensuring efficient resource utilization, and minimizing delays. leading to the funding of various projects by SESAR (Single European Sky ATM Research). Among these, the TAPAS specializes in air traffic flow, capacity management, and air traffic control services within ATM. The project's scope includes establishing transparency requirements, developing a prototype, implementing an explainable layer, and formulating principles for transparency in ATM automation processes, as referenced in \cite{tapas}. ARTIMATION improves runway delay predictions and optimizes departures and inspections by enhancing explainability with heatmaps and decision-based storytelling. Additionally, the Mindtooth tool evaluates air traffic controllers' reactions, including Approach-Withdrawal, stress, and workload using neurophysiological metrics. This is particularly relevant considering that time pressure is a critical factor in high-stakes environments like air traffic control, where rapid and accurate decision-making is imperative \cite{artimation2024measuring}.

Concerning Demand-Capacity Balancing (DCB), challenges such as congestion or hotspots occur when airspace demand exceeds capacity. Research in this area includes a multi-agent RL approach that uses a rule-based environment and a modified proximal policy optimization (MAPPO) framework. This approach integrates a supervisor for enhanced learning, effectively balancing demand, and capacity in air traffic flow management \cite{tang2021multiagent}. In \cite{kravaris2023explaining}, EI enhances the explainability of deep multi-agent reinforcement learning solutions for demand-capacity balance (DCB) problems, employing visual analytics for transparent decision-making. Additionally, XAI can drive secure data chains in conjunction with blockchain networks for autonomous ATM, ensuring data integrity in ATM operations \cite{axon2023securing}. 

\subsubsection{Challenges for Explainable Interface in Transportation}
Summarizing key findings, integrating XAI into transportation systems is crucial for high safety and operational standards. Major challenges include:

\begin{itemize}
  \item Integration with Existing Infrastructure: Merging XAI with existing transportation infrastructure, such as traffic signals and GPS systems.
  \item Adapting to Dynamic Environments such as unpredictable traffic, weather, and pedestrian behavior.
  \item Real-Time Data Processing and Interaction: In high-stake real-time HAT, swiftly processing dynamic data, provide understandable explanations in proper time and adjust EIs according to feedback.
  \item Real-time Evaluation:  Collecting evaluation data during HAT process, without distracting the user.
  \item Challenges in Advanced Urban Transportation: The emergence of intricate urban air traffic, featuring drones and eVTOLs (electric vertical take-off and landing vehicles), presents unique challenges and prospects for incorporating EIs in AVs and traffic control systems. Addressing these issues calls for creative solutions to preserve both efficiency and safety within complex urban transportation networks.
\end{itemize}

\subsection{Explainable Interface and Digital Twin}
\subsubsection{Relationship of Physical System, Digital Twin, and XAI}
The relationship between a physical system and its digital twin (DT) is dynamic and symbiotic. The DT mirrors the life of its corresponding physical entity, adapting its real-time behavior in response to inputs from the physical system. In the context of Industry 4.0, Cyber-Physical Systems (CPS) provide real-time data and control capabilities \cite{groshev2021intelligent}. Utilizing DT for decision-making, they enable the simulation and modelling of various aspects of a physical system, offering valuable insights and predictions. These simulations can serve multiple purposes, such as optimizing performance, predicting maintenance needs, and simulating potential scenarios in smart cities, manufacturing, healthcare \cite{fuller2020digital}. EIs play a crucial role in this relationship by ensuring transparency and understandability in the decision-making processes within the DT.
\subsubsection{XAI Methods for Digital Twin}
XAI methods play a pivotal role in DT. For example, LSTM neural networks, employed for lane change prediction in autonomous driving DT, use techniques such as layer-wise relevance propagation for better clarity in their predictions \cite{wehner2022explainable}. Furthermore, architectures like TwinExplainer integrate XAI into deep learning-powered DT systems, prioritizing the interpretability of their predictive outcomes \cite{neupane2023twinexplainer}. The study \cite{kobayashi2023explainable} investigated the remaining useful life prediction in DT systems through interpretable machine learning, employing the PiML library for transparent and reliable AI decision-making.
\subsubsection{Digital Twin as an Explainable Interface}
DTs surpass traditional explainable interfaces (EIs) by offering interactive features and counterfactual analysis. This improves understanding of AI-driven decisions across various industries, like healthcare and manufacturing. For instance, In drug discovery, DTs explain AI's reasoning behind diagnoses and treatments, boosting trust and confident decision-making by healthcare professionals. \cite{askr2023deep}. In the manufacturing industry, DTs can provide real-time explanations for quality control decisions, helping operators identify the root causes of defects and optimize production processes \cite{park2019challenges}.
\subsection{Challenges for XAI and Digital Twin Integration}
The integration of XAI methodologies into DT systems presents interface design and optimization challenges \cite{javed2023survey}. First, incorporating XAI not only as a post-hoc approach to explain the outputs, but also integrating internal interpretable AI from the initial stages of design. Moreover, the progression of DT technology is hindered by a general lack of understanding of its advantages and a deficiency in technical expertise. Contributing technologies such as 3D simulations, IoT, AI, big data, and cloud computing are also in developmental stages, which adds to the complexity \cite{tao2018digital}. Furthermore, advancements in high-performance computing and real-time interaction technologies are crucial for the enhancement of XAI-enabled DTs. This necessitates a high level of technical understanding from users. To address these issues, there is a need for the development of new methods and frameworks. These should enable the seamless integration of XAI into DT systems, thus lowering the learning curve for users and making it easier for them to understand and trust these systems.

\end{document}